\documentclass[sigconf]{acmart}

\AtBeginDocument{%
  \providecommand\BibTeX{{%
    \normalfont B\kern-0.5em{\scshape i\kern-0.25em b}\kern-0.8em\TeX}}}

\renewcommand\footnotetextcopyrightpermission[1]{}
\usepackage[ruled]{algorithm2e}
\usepackage{diagbox}
\usepackage{makecell}
\usepackage{graphicx}
\usepackage{multirow}
\usepackage{colortbl}
\usepackage{hyperref}
\usepackage[normalem]{ulem}
\useunder{\uline}{\ul}{}

\begin{document}

\title{Probing Unlearned Diffusion Models: \\ A Transferable Adversarial Attack Perspective}



\author{Xiaoxuan Han}
\affiliation{%
  \institution{University of Chinese Academy of Sciences}
  \city{Beijing}
  \country{China}}
\email{hanxiaoxuan2023@ia.ac.cn}

\author{Songlin Yang}
\affiliation{%
  \institution{University of Chinese Academy of Sciences}
  \city{Beijing}
  \country{China}}
\email{yangsonglin2021@ia.ac.cn}

\author{Wei Wang}
\authornote{Corresponding author.}
\affiliation{%
  \institution{Institute of Automation, Chinese Academy of Sciences}
  \city{Beijing}
  \country{China}}
\email{wwang@nlpr.ia.ac.cn}

\author{Yang Li}
\affiliation{%
  \institution{University of Chinese Academy of Sciences}
  \city{Beijing}
  \country{China}}
\email{liyang2022@ia.ac.cn}

\author{Jing Dong}
\affiliation{%
  \institution{Institute of Automation, Chinese Academy of Sciences}
  \city{Beijing}
  \country{China}}
\email{jdong@nlpr.ia.ac.cn}



\begin{abstract}

  Advanced text-to-image diffusion models raise safety concerns regarding identity privacy violation, copyright infringement, and Not Safe For Work (NSFW) content generation (e.g., nudity). Towards this, unlearning methods have been developed to erase these involved concepts from diffusion models. However, these unlearning methods only shift the text-to-image mapping and preserve the visual content within the generative space of diffusion models, leaving a fatal flaw for restoring these erased concepts. This erasure trustworthiness problem needs probe, but previous methods are sub-optimal from two perspectives: \textbf{(1) Lack of transferability}: Some methods operate within a white-box setting, requiring access to the unlearned model. And the learned adversarial input often fails to transfer to other unlearned models for concept restoration; \textbf{(2) Limited attack}: The prompt-level methods struggle to restore narrow concepts from unlearned models, such as celebrity identity. Therefore, this paper aims to leverage the transferability of the adversarial attack to probe the unlearning robustness under a black-box setting. This challenging scenario assumes that the unlearning method is unknown and the unlearned model is inaccessible for optimization, requiring the attack to be capable of transferring across different unlearned models. Specifically, we first analyze the reasons for the poor transferability of previous methods. Then, we employ an adversarial search strategy to search for the adversarial embedding which can transfer across different unlearned models. This strategy adopts the original Stable Diffusion model as a surrogate model to iteratively erase and search for embeddings, enabling it to find the embedding that can restore the target concept for different unlearning methods. Extensive experiments demonstrate the transferability of the searched adversarial embedding across several state-of-the-art unlearning methods and its effectiveness for different levels of concepts. Our code is available at \href{https://github.com/hxxdtd/PUND}{https://github.com/hxxdtd/PUND}. \textbf{\textit{CAUTION: This paper contains model-generated content that may be offensive}}.
  
\end{abstract}


\begin{teaserfigure}
    \centering
  \includegraphics[width=0.95\textwidth]{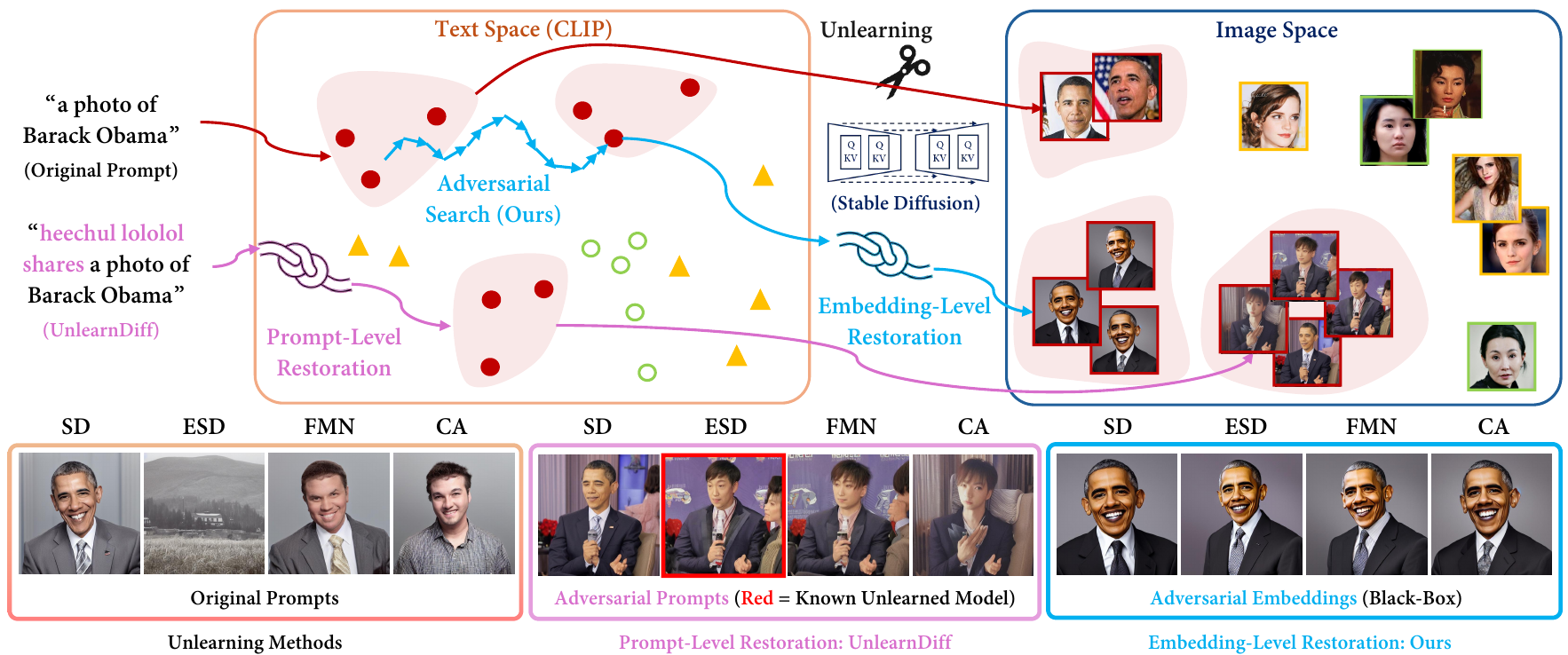}
  \caption{The overview of concept erasure and restoration for text-to-image Stable Diffusion~\cite{latent_diffusion} (SD) model. Due to content safety concerns, unlearning methods (e.g., ESD~\cite{esd}, FMN~\cite{fmn}, and CA~\cite{ca}) have been investigated to erase target concepts by shifting the text-to-image mapping. But these erasure methods leave a fatal flaw for restoring the erased concepts. Towards this, we propose an adversarial attack to probe erasure trustworthiness. Compared with previous methods, our black-box method can transfer across different unlearned models and is effective for the challenging task of celebrity identity restoration.}
  \label{fig:teaser}
\end{teaserfigure}



\maketitle

\section{Introduction}

Developing Artificial Intelligence Generated Content (AIGC) is a double-edged sword. Although text-to-image (T2I) generative models \cite{latent_diffusion, imagen, dalle, glide, vqdiffusion, parti} can generate high-quality and diverse images according to the given prompts, they also raise significant safety concerns regarding identity privacy \cite{extract}, copyright \cite{art_forgery}, and Not Safe For Work (NSFW) content \cite{sld, redteam}. For example, these models can generate portraits of celebrities known as deepfakes \cite{deepfake_diffusion}, while some painting styles of artists can be imitated. Besides, these models can generate Not Safe For Work (NSFW) content, such as nudity and violence. To mitigate these issues, as shown in Figure ~\ref{fig:teaser}, concept erasure methods \cite{uce, esd, fmn, ca, sa, salun} for erasing the involved concepts have been developed, falling under the category of machine unlearning \cite{mu1, mu2, mu3, mu_fast, mu_sparse}. However, existing methods accomplish the ``erasure'' task by shifting the text-to-image mapping and fail to erase the visual content within the generative space of diffusion models, leaving a fatal flaw for restoring these erased concepts. This inspires the question: \textbf{How trustworthy are unlearning methods for text-to-image diffusion models?}

Previous methods are sub-optimal for this question from two perspectives. \textbf{(1) Lack of transferability}: Some methods operate within a white-box setting, requiring access to the unlearned model. And the learned adversarial input often fails to transfer to other unlearned models for concept restoration. As shown in Figure ~\ref{fig:teaser}, methods like UnlearnDiff~\cite{unlearndiff} make a strong assumption that the unlearned generative model is accessible, which lacks transferability and is less practical in real-world scenarios. \textbf{(2) Limited attack}: The prompt-level methods struggle to restore narrow concepts from unlearned models, such as celebrity identity (ID). As depicted in Figure ~\ref{fig:teaser}, UnlearnDiff~\cite{unlearndiff} tries to optimize an adversarial prompt to restore the target concept (i.e., ``Barack Obama''). But its representative capability is constrained by the discrete nature of prompt tokens, making accurate identity restoration challenging.

In this paper, we aim to probe the erasure robustness of unlearned diffusion models under the black-box setting, where the attacker lacks knowledge of the unlearning methods and the unlearned models are inaccessible for the optimization. This is significantly challenging, especially for narrow concepts such as identity. We first analyze the reasons for the poor transferability of previous method. To tackle this challenge, we utilize an adversarial search strategy to find the adversarial embedding transferable across different unlearned models. This strategy adopts an original Stable Diffusion model as a surrogate model to alternately erase and search for embeddings, guiding the embedding search from high-density regions to low-density ones. These embeddings located in low-density regions are difficult to erase, enabling them to restore the target concept for different unlearned methods.

\textbf{Our contributions are summarized as follows:}
\begin{itemize}
    
    \item We propose a transferable adversarial attack to probe the unlearning robustness, which can transfer across diverse unlearned models and tackle the challenge of ID restoration.
    \item We improve the transferability by iteratively erasing and searching for the embeddings that can restore the target concept. The obtained embeddings are located in low-density regions and very likely to be overlooked by erasure methods, thus possessing greater restoration capabilities.
    \item Extensive experiments demonstrate the transferability of the searched adversarial embedding across various state-of-the-art unlearning methods, along with its effectiveness across diverse levels of concepts ranging from broad to narrow.
\end{itemize}

\section{Related Work}
\subsection{Text-to-Image Diffusion Models}
Text-to-Image (T2I) Diffusion models are a variant of diffusion models, providing text-conditional guidance for image generation. The training of diffusion models includes two processes \cite{ddpm, ddim}. In the diffusion process, noise is gradually added to the image $x_0$ over multiple steps. In the reverse process~\cite{dpmsolver, ddim, pndm}, the model learns to predict the noise given the time step $t$ and the noised version image $x_{t}$. T2I diffusion models \cite{latent_diffusion, vqdiffusion, dalle} typically conduct the diffusion and reverse process in the latent space~\cite{vqgan} of lower dimension for better computation efficiency. And various conditional mechanisms~\cite{controlnet, ip-adapter, latent_diffusion} (such as text prompt), are introduced for diverse controllable T2I generation. But the inappropriate text input can result in undesirable generated images~\cite{yang2024position, sld, zhai2024discovering}, prompting the development of concept erasure methods to address this issue.
\vspace{-0.3cm}

\begin{figure*}[]
\centering
\includegraphics[width=0.95\textwidth]{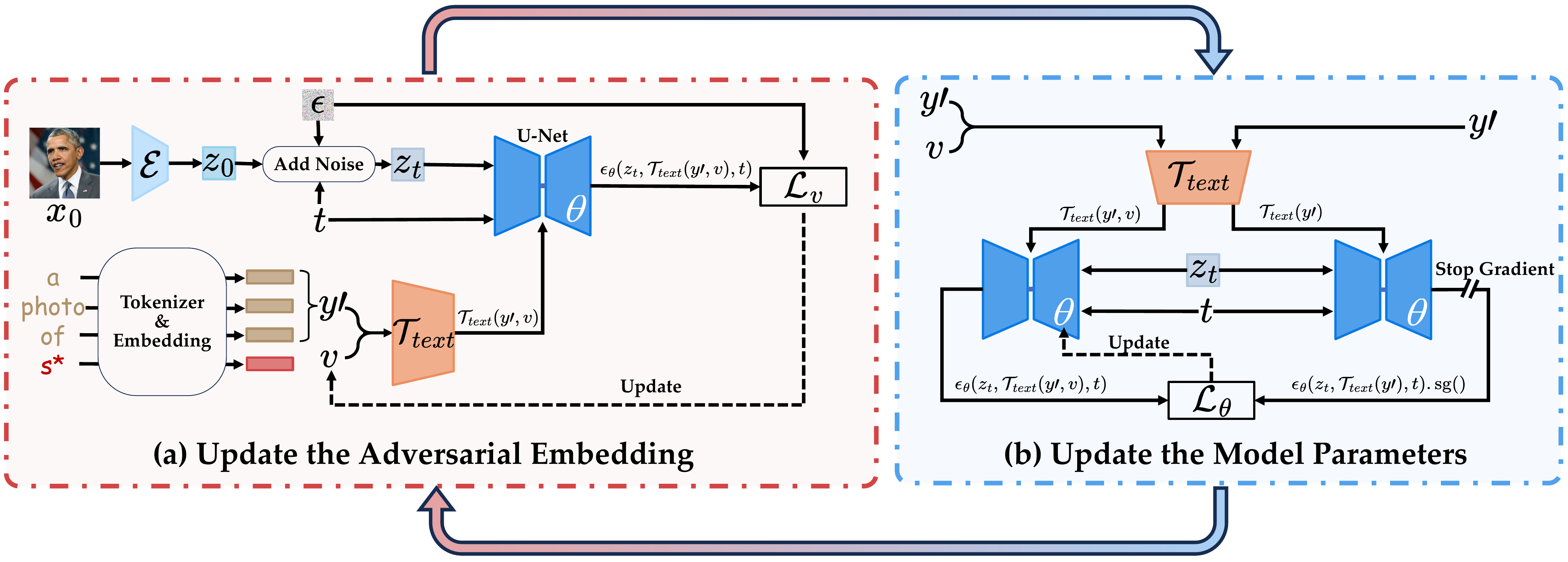}
\caption{The Adversarial Search (AS) strategy for concept restoration. We adopt the original Stable Diffusion model as a surrogate model to alternately erase and search for embeddings which can restore the target concepts.}
\label{pipeline}
\vspace{-0.2cm}
\end{figure*}

\subsection{Diffusion Unlearning for Concept Erasure}

A direct way to remove undesirable concepts is to filter out the inappropriate images and use the remaining data to retrain the model. But it is hard to find a perfect classifier to detect inappropriate images, and retraining the model from scratch requires large amounts of computation resources. Therefore, existing unlearning methods focus on erasing concepts from the trained model. Erased Stable Diffusion (ESD) \cite{esd} guides the predicted noise in the opposite direction of the target concept (i.e., the concept to erase) to decrease the probability of target concept generation. Concept Ablation (CA) guides the distribution of target concept towards a broader concept called the anchor concept. Forget-Me-Not (FMN) \cite{fmn} doesn't direct the target concept distribution to a specified distribution; instead, it minimizes the attention map corresponding to the target concept. Unlike the above fine-tuning-based methods, Unified Concept Editing (UCE) \cite{uce} is a model editing method that applies a closed-form editing to the liner projection weights of cross-attention parts. While existing unlearning methods demonstrate good performance under benign input, their reliability under adversarial scenarios requires further investigation. 
\vspace{-0.3cm}

\subsection{Adversarial Concept Restoration}
UnlearnDiff attack \cite{unlearndiff} uses the classifier nature of diffusion models and employs Projected Gradient Descent (PGD) to optimize in the token space for concept restoration. But it requires access to the unlearned model. In contrast, Ring-A-Bell \cite{ringabell} assumes that the unlearned model is inaccessible and utilizes the text encoder to search for adversarial prompts. It employs prompts with and without the target concept to obtain its representation in the embedding space, then utilizes a genetic algorithm to search for adversarial prompts in the discrete token space. While Ring-A-Bell mainly focuses on NSFW content, it overlooks the probe of narrower concepts like celebrity identity. Pham et al. \cite{circumventing} probe a wide range of concepts, including celebrity identity, artist styles and NSFW content, with the aid of Textual Inversion \cite{textual_inversion}. However, they also require access to the unlearned model like \cite{unlearndiff}. To overcome the limitations of existing methods, we exploit the transferability of adversarial embedding to probe the restoration of diverse concepts, particularly narrow ones, without needing access to the unlearned model during the optimization process.

\section{Method}

In this section, we first introduce the preliminaries for the Stable Diffusion model, and explain how to conduct concept restoration at the embedding level. Then, we analyze the reasons for the poor transferability of the naive method and present the motivation behind the proposed strategy for improving transferability. Finally, we introduce the formulation and implementation of this strategy.

\subsection{Preliminaries}
\textbf{Stable Diffusion.} We focus on the widely employed latent diffusion models \cite{latent_diffusion}, also known as Stable Diffusion. In the training process, the image $x$ undergoes encoding through the encoder $\mathcal{E}$, yielding its latent representation $z=\mathcal{E}(x)$. Subsequently, Gaussian noise is gradually added to $z_0$, generating a sequence of noised versions of the latent representation \{$z_1$, $z_2$, $\dots$, $z_T$\}, where $T$ denotes the total number of time steps. The training objective is to predict the Gaussian noise $\epsilon$ as follows:

\begin{equation}
    \mathcal{L}=\mathbb{E}_{z\sim\mathcal{E}(x),y,\epsilon\sim\mathcal{N}(0,1),t}\Big[\|\epsilon-\epsilon_\theta(z_t,\mathcal{T}_{text}(y),t)\|_2^2\Big],
\end{equation}
where $y$ represents the token embeddings of the input prompt, $\mathcal{T}_{text}$ denotes the text encoder used to obtain the conditional guidance of $y$, $t$ is the current time step, and $\theta$ denotes model parameters.

\noindent
\textbf{Concept Restoration.} To restore the target concept from the unlearned model parameterized by $\hat{\theta}$, Pham et al. \cite{circumventing} utilized Textual Inversion \cite{textual_inversion} to obtain the adversarial embedding for restoration. Concretely, they replaced the target concept in the input prompt with the placeholder token $S^{*}$ (e.g., \textit{"a photo of <target-concept>"} is modified as \textit{"a photo of $S^{*}$"}), and optimized the corresponding embedding of $S^{*}$ denoted by $v$. The modified prompt is then fed into the text encoder to obtain the new text condition $\mathcal{T}_{text}(y\prime,v)$, where $y\prime$ denotes the token embeddings of the original prompt but without the target concept (e.g., \textit{"a photo of"}). Pham et al. \cite{circumventing} applied Textual Inversion to the unlearned model to find the optimal embedding for the token $S^{*}$ by solving the optimization problem below:

\begin{equation}
    v^*=\arg\min_{v}\mathbb{E}_{z\sim\mathcal{E}(x),y\prime,\epsilon\sim\mathcal{N}(0,1),t}\Big[\|\epsilon-\epsilon_{\hat{\theta}}(z_t,\mathcal{T}_{text}(y\prime,v),t)\|_2^2\Big]
\end{equation}
where $v^{*}$ denotes the learned adversarial embedding. 

The method proposed by Pham et al.~\cite{circumventing} proves effective for restoring diverse concepts, including narrow ones such as celebrity identity, which are challenging to restore using prompt-level attack methods such as Ring-A-Bell~\cite{ringabell} and UnlearnDiff~\cite{unlearndiff} due to the discrete nature of text representation. However, Pham et al.~\cite{circumventing} made a strong assumption that the attacker had access to unlearned models. In this paper, we aim to perform concept restoration with access only to the original Stable Diffusion model (i.e., the model before unlearning) parameterized by $\theta$, and leverage the transferability of the adversarial embedding to probe the robustness of different unlearning methods.

\begin{figure}
\centering
\includegraphics[width=0.4\textwidth]{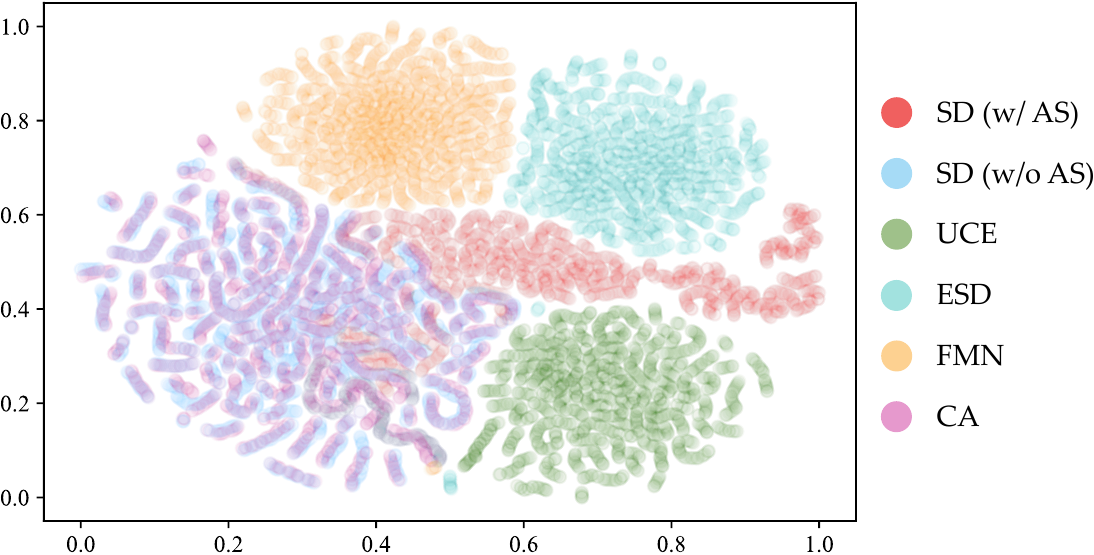}
\caption{The visualization of the embeddings obtained from different unlearned models for the restoration of ``Barack Obama''. Purple, yellow, cyan, and green points represent embeddings obtained from the models that have been unlearned by CA~\cite{ca}, FMN~\cite{fmn}, ESD~\cite{esd}, and UCE~\cite{uce}, respectively. The blue and red points represent the embeddings acquired from the original Stable Diffusion (SD) model, while the red ones are obtained with our Adversarial Search (AS) strategy.}
\vspace{-0.3cm}
\label{figure_tsne}
\end{figure}

\subsection{Transferable Adversarial Search Strategy}

\noindent\textbf{Analysis for Poor Transferability.} Concept restoration aims to find the embedding capable of restoring the target concept within the model. A naive approach involves directly applying Textual Inversion \cite{textual_inversion} to the original model. However, as illustrated in Figure~\ref{figure_tsne}, the scatter plot displays that embeddings, represented by cyan, blue, yellow, purple, and green points, obtained from a single model—whether it is the original or an unlearned one—tend to exhibit distinctly separated cluster distributions. In other words, these embeddings obtained from the original model are highly likely to be ineffective for unlearned models. One possible explanation is that, for the original model, the learned embedding closely resembles the token embedding of the target concept, thus having a high probability of being erased. The transferable embedding we seek needs to diverge from the token embedding of the target concept. However, the embedding optimized for the original model may be stuck around the token embedding of the target concept due to the local minimum nature inherent in gradient-based optimization methods. Hence, additional guidance is necessary to facilitate the optimization process and explore more potential regions.

\noindent\textbf{How to Improve Transferability?} In the continuous embedding space, the distribution of the target concept may conform to a specific distribution, with high-density regions likely centered around the token embedding of the target concept. Given that most unlearning methods \cite{esd,uce,ca} utilize the token of the target concept for erasure, these high-density regions are mostly erased. Consequently, the transferable embedding for target concept restoration is likely to reside in low-density regions. Thus we need to guide the embedding search from high-density regions to low-density ones. To achieve this, we employ an Adversarial Search (AS) strategy. This strategy initially seeks the embedding from the original model, attempts to erase it, and then repeats the process of embedding search and erasure iteratively. After a number of iterations, we can obtain a series of embeddings, among which lies the transferable one. As depicted in Figure \ref{figure_tsne} by the red points, after applying our adversarial search strategy, it is possible to avoid falling into the distribution of a single model. Therefore, the red portion exhibits a bar-like distribution, showing a tendency to intersect with all distributions where restorations are successful.

\begin{algorithm}
\caption{Searching for the transferable embedding for concept restoration}
\label{algorithm}
\LinesNumbered
\KwIn{Original model parameterized by $\theta$, training images $X$, total epochs $E$, image encoder $\mathcal{E}$, embedding update iterations per epoch $I_{v}$, total time steps $T$, noise coefficient sequence $(\overline{\alpha}_t)_{t=1}^T$, text encoder $\mathcal{T}_{text}$, neutral token embeddings $y\prime$, parameter update frequency $f$, parameter update iterations $I_{\theta}$}
\KwOut{Candidate embedding set $V$}
Initialize: $v\leftarrow v_0$\;
\For{\rm{$e$ in range($E$)}}{
    $x_0 \in X$; \tcp{randomly pick the training image}
    $z_0=\mathcal{E}(x_0)$ \;
    \tcp{update the embedding for $I_{v}$ iterations}
    \For{\rm{$i$ in range($I_{v}$)}}
    {
        $t\sim\mathrm{Uniform}([1...T]),\epsilon\sim\mathcal{N}(0,I)$ \;
        $z_t=\sqrt{\overline{\alpha}_t}z_0+\sqrt{1-\overline{\alpha}_t}\epsilon$ \;
        $\mathcal{L}_{v}=\|\epsilon-\epsilon_{\theta}(z_{t},{\mathcal T}_{text}(y\prime,v),t)\|_{2}^{2}$ \;
        \text{Update $v$ with gradient descent to minimize $\mathcal{L}_{v}$} \;
    }
    $V \leftarrow V \cup \{v\}$; \tcp{add $v$ to the candidate set}
    \tcp{update model parameters every $f$ epochs}
    \If{$f \mid e$}{
        \For{\rm{$i$ in range($I_{\theta}$)}}
        {
            $t\sim\mathrm{Uniform}([1...T]),\epsilon\sim\mathcal{N}(0,I)$ \;
            $z_t=\sqrt{\overline{\alpha}_t}z_0+\sqrt{1-\overline{\alpha}_t}\epsilon$ \;
            $\mathcal{L}_{\theta}=\|\epsilon_{\theta}(z_t,\mathcal{T}_{text}(y\prime,v),t)-\epsilon_{\theta}(z_t,\mathcal{T}_{text}(y\prime),t).\mathrm{sg}()\|_2^2$ \;
            \text{Update $\theta$ with gradient descent to minimize $\mathcal{L}_{\theta}$};
        }
    }
}
\textbf{return} $V$
\end{algorithm}

\noindent\textbf{Strategy Formulation.} Adversarial search is formulated as a Min-Max optimization process, expressed as follows:

\begin{equation}
    \min_{v}\max_{\theta}\mathbb{E}_{z\sim\mathcal{E}(x),y\prime,\epsilon\sim\mathcal{N}(0,1),t}\Big[\|\epsilon-\epsilon_{\theta}(z_t,\mathcal{T}_{text}(y\prime,v),t)\|_2^2\Big].
\end{equation}
The inner maximization aims to update the model parameters to maximize the noise prediction loss, preventing the learned embedding from restoring the target concept. However, directly maximizing the loss can significantly impair the model's capacity in image generation, as also noted in \cite{ca}. Thus, we relax the inner maximization by introducing $\tilde{\epsilon}$, satisfying $\|\epsilon-\tilde{\epsilon}\|_2\geq d>0$, where $d$ is a constant. In the inner maximization, maximizing $\|\epsilon-\epsilon_{\theta}(z_t,\mathcal{T}_{text}(y\prime,v),t)\|_2^2$ is equivalent to maximizing $\|\epsilon-\epsilon_{\theta}(z_t,\mathcal{T}_{text}(y\prime,v),t)\|_2$, which can be rewritten as follows:

\begin{align}
        \! \big\|\epsilon-\epsilon_{\theta}(z_t,\mathcal{T}_{text}(y\prime,v),t)\big\|_2 &= 
        \big\|\big(\epsilon-\tilde{\epsilon}\big)-\big(\epsilon_{\theta}(z_t,\mathcal{T}_{text}(y\prime,v),t)-\tilde{\epsilon}\big)\big\|_2 \nonumber \\
        &\geq \big\|\epsilon-\tilde{\epsilon}\big\|_2 - \big\|\epsilon_{\theta}(z_t,\mathcal{T}_{text}(y\prime,v),t)-\tilde{\epsilon}\big\|_2 \nonumber \\
        &\geq d - \big\|\epsilon_{\theta}(z_t,\mathcal{T}_{text}(y\prime,v),t)-\tilde{\epsilon}\big\|_2. \!
\end{align}

Instead of maximizing the loss itself, we relax it by maximizing its lower bound, which is equivalent to a minimization problem as below:

\begin{equation}
    \min_{\theta} \big\|\epsilon_{\theta}(z_t,\mathcal{T}_{text}(y\prime,v),t)-\tilde{\epsilon}\big\|_2^2.
\end{equation}

Then we need to determine the assignment of $\tilde{\epsilon}$. Naturally, $\tilde{\epsilon}$ should maintain a certain distance $d$ from $\epsilon$ to ensure a large lower bound. In practice, we adopt the following assignment:

\begin{equation}
    \Tilde{\epsilon}=\epsilon_{\theta_0}(z_t,\mathcal{T}_{text}(y\prime),t),
\end{equation}
where $\theta_0$ is the original model parameters (i.e., parameters before updating), $y\prime$ is token embeddings of a neutral prompt (i.e., \textit{"a photo of"}). Then, the inner optimization becomes:

\begin{equation}
    \min_{\theta} \big\|\epsilon_{\theta}(z_t,\mathcal{T}_{text}(y\prime,v),t)-\epsilon_{\theta_0}(z_t,\mathcal{T}_{text}(y\prime),t)\big\|_2^2.
\end{equation}
But the above minimization process has a relatively large memory burden, as the attacker needs to make a copy of the model parameters (i.e., $\theta_0$) before updating. To address this, we use $\epsilon_{\theta}(z_t,\mathcal{T}_{text}(y\prime),t)$ to approximate $\epsilon_{\theta_0}(z_t,\mathcal{T}_{text}(y\prime),t)$, assuming that the model maintains the capacity to generate neutral images during parameter updating, which is similar to the assumption in \cite{ca}. Then the minimization process can be written as:

\begin{equation}
    \min_{\theta} \big\|\epsilon_{\theta}(z_t,\mathcal{T}_{text}(y\prime,v),t)-\epsilon_{\theta}(z_t,\mathcal{T}_{text}(y\prime),t).\mathrm{sg}()\big\|_2^2,
\end{equation}
where $.\mathrm{sg}()$ represents the stop gradient operation, ensuring that the model's ability to generate neutral images is not damaged. This minimization process remaps the learned embedding to neutral images, thereby removing the mapping from the embedding to the target concept. It's important to note that this erasing formulation differs from existing methods. While CA \cite{ca} appears to perform erasing in a similar manner, it needs to determine an anchor concept, making it not as general as the approach used here.

\noindent\textbf{Strategy Implementation.} Then we introduce the process of searching for the transferable embedding for concept restoration, as presented in Figure~\ref{pipeline}. The attacker begins by collecting some images of the target concept, denoted by $X=\{x^{i}\}_{i=1}^{N}$, where $N$ is the total number of images. During the optimization of the embedding, the reference image $x_0$ is randomly sampled from $X$ and input to the image encoder to obtain its latent representation $z_0=\mathcal{E}(x_0)$. The time step $t$ is uniformly sampled from $1$ to $T$, where $T$ is the total number of time steps. The noise $\epsilon$ is randomly sampled from a Gaussian distribution. Next, the noised version of $z$ at step $t$ can be calculated as: $z_t=\sqrt{\overline{\alpha}_t}z_0+\sqrt{1-\overline{\alpha}_t}\epsilon$, where $\overline{\alpha}_t$ is a predefined constant for noise addition. The noise prediction loss can be computed as: $\mathcal{L}_{v}=\|\epsilon-\epsilon_{\theta}(z_{t},{\mathcal T}_{text}(y\prime,v),t)\|_{2}^{2}$. Then the embedding is updated using gradient descent to minimize $\mathcal{L}_{v}$ while keeping the model parameters $\theta$ fixed. When updating the model parameters $\theta$, the loss is calculated as: $\|\epsilon_{\theta}(z_t,\mathcal{T}_{text}(y\prime,v),t)-\epsilon_{\theta}(z_t,\mathcal{T}_{text}(y\prime),t).\mathrm{sg}()\|_2^2$. Gradient descent is then applied to update $\theta$ while keeping $v$ fixed. The detailed process of the strategy can be found in Algorithm \ref{algorithm}.

\begin{figure*}
\centering
\includegraphics[width=0.95\textwidth]{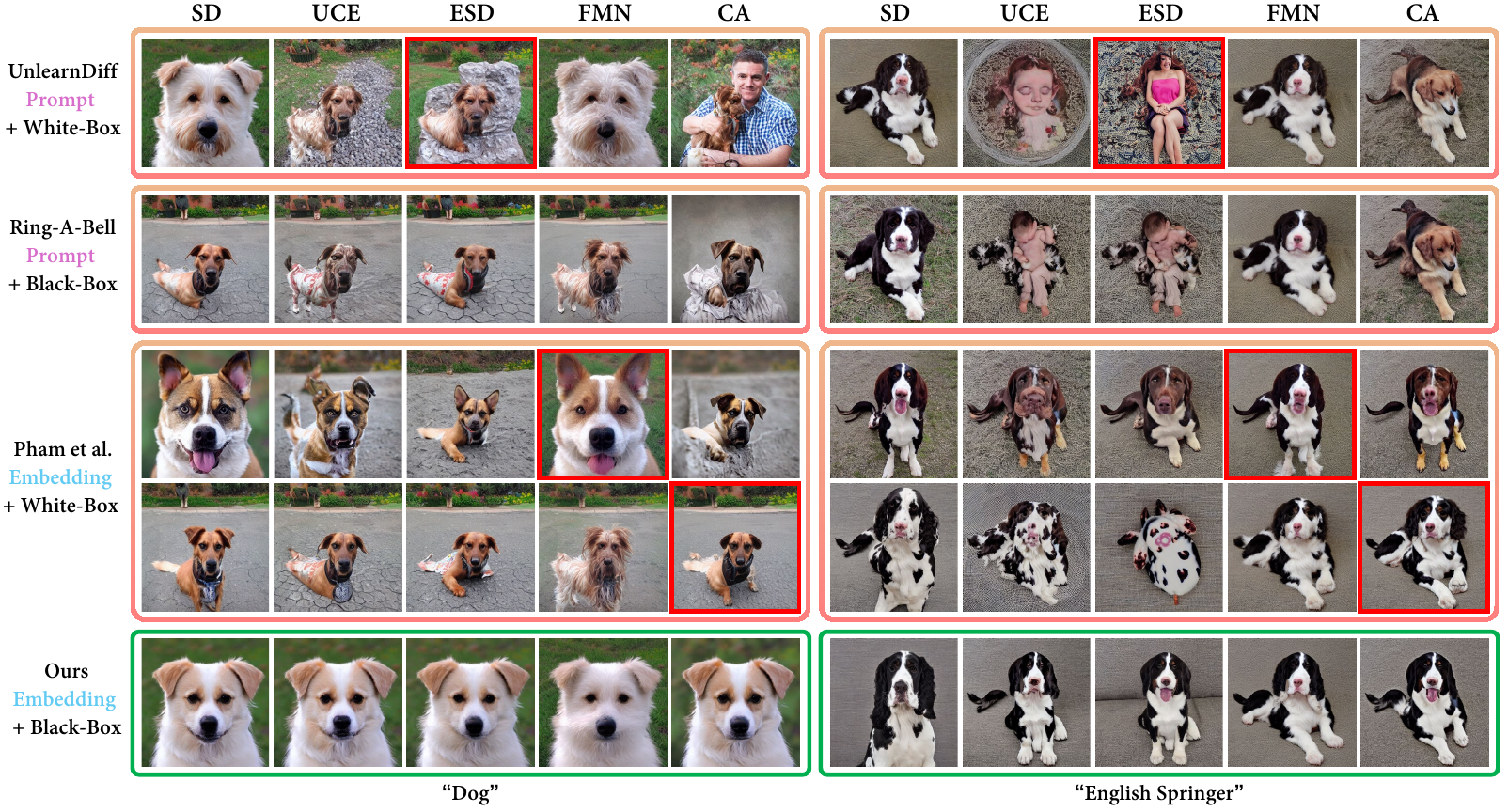}
\caption{The comparisons with different concept restoration methods for objects, encompassing both broad and narrow objects.}
\label{figure_object}
\vspace{-0.3cm}
\end{figure*}

\begin{figure*}
\centering
\includegraphics[width=0.95\textwidth]{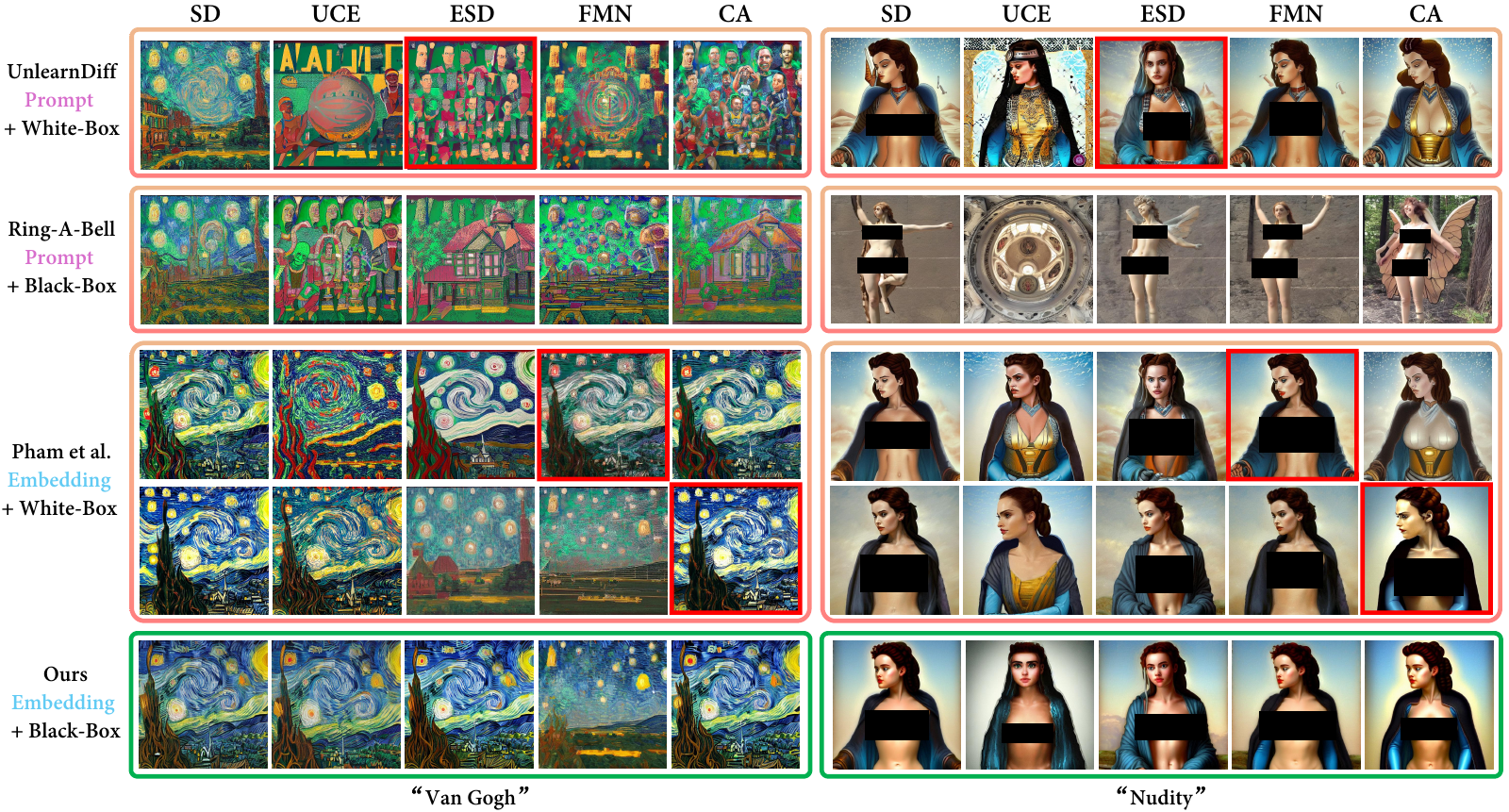}
\caption{The comparisons with different concept restoration methods for artist styles and NSFW content.}
\label{figure_nsfw+style}
\end{figure*}

\begin{figure*}
\centering
\includegraphics[width=0.95\textwidth]{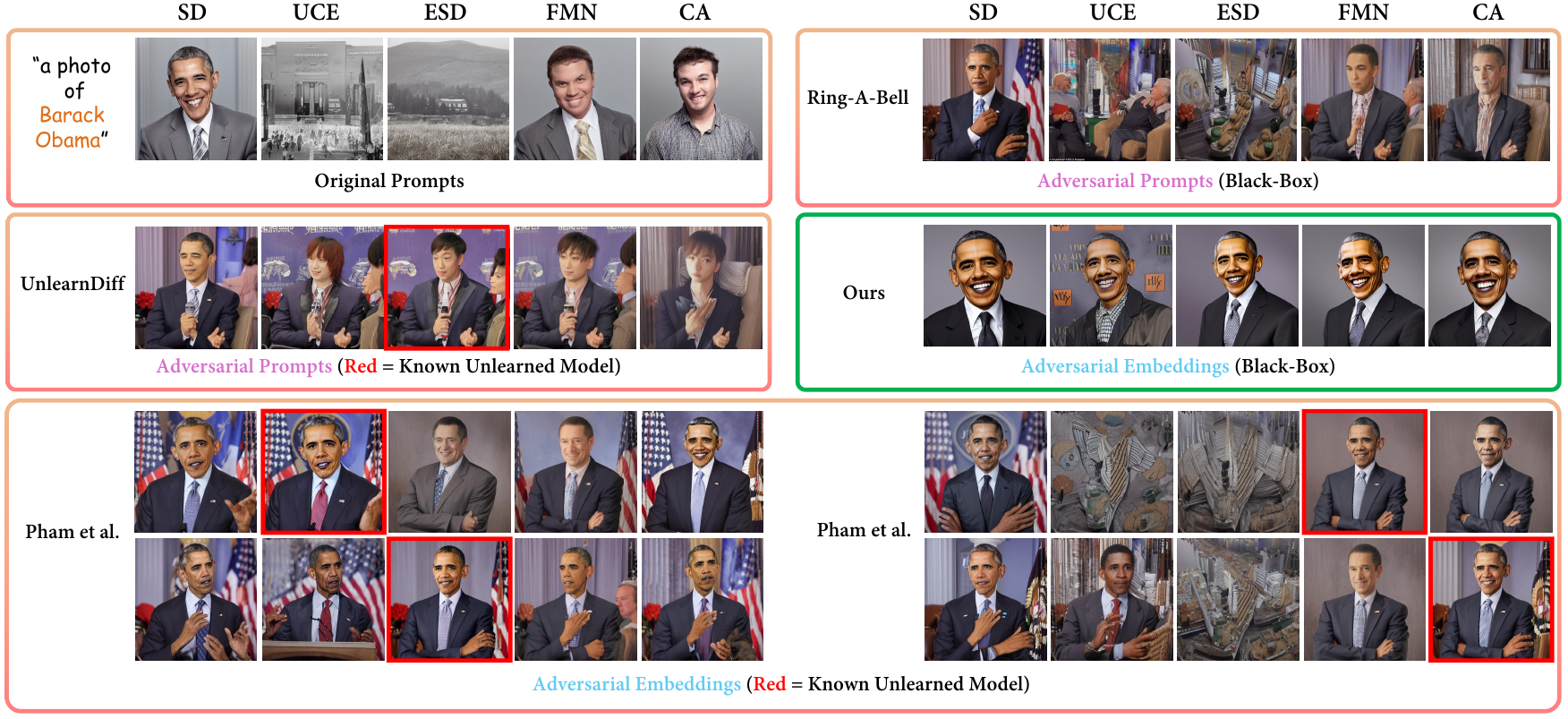}
\caption{The comparisons with different concept restoration methods for identities.}
\label{figure_id}
\end{figure*}

\begin{table}[]
\caption{Comparisons of Different Attack Methods on Diverse Concepts. The best results in \textbf{bold}, the second best \underline{underlined}. The asterisk (*) denotes a white-box attack.}
\label{table_cmp}
\footnotesize
\begin{tabular}{c|c|cccccc}
\Xhline{1.0pt}
 &  & \multicolumn{6}{c}{\textbf{Attack Methods}} \\ \cline{3-8} 
\multirow{-2}{*}{\textbf{\begin{tabular}[c]{@{}c@{}}Target\\ Concepts\end{tabular}}} & \multirow{-2}{*}{\textbf{\begin{tabular}[c]{@{}c@{}}Erasure\\ Methods\end{tabular}}} & \multicolumn{1}{c|}{\begin{tabular}[c]{@{}c@{}}w/o \\ Attack\end{tabular}} & UD & RAB & \begin{tabular}[c]{@{}c@{}}CI\\ (FMN)\end{tabular} & \multicolumn{1}{c|}{\begin{tabular}[c]{@{}c@{}}CI\\ (CA)\end{tabular}} & Ours \\ \Xhline{1.0pt}
 & UCE & \multicolumn{1}{c|}{0.0} & 37.0 & 29.0 & {\ul 86.0} & \multicolumn{1}{c|}{82.0} & \textbf{96.0} \\
 & ESD & \multicolumn{1}{c|}{16.0} & 57.0* & 62.0 & {\ul 93.0} & \multicolumn{1}{c|}{78.0} & \textbf{98.0} \\
 & FMN & \multicolumn{1}{c|}{65.0} & 64.0 & 66.0 & {\ul 82.0*} & \multicolumn{1}{c|}{74.0} & \textbf{84.0} \\
 & CA & \multicolumn{1}{c|}{12.0} & 62.0 & 59.0 & 90.0 & \multicolumn{1}{c|}{{\ul 98.0*}} & \textbf{100.0} \\
\multirow{-5}{*}{\begin{tabular}[c]{@{}c@{}}Dog\\ (Object)\end{tabular}} & \cellcolor[HTML]{FFC3C3}Average & \multicolumn{1}{c|}{\cellcolor[HTML]{FFC3C3}23.3} & \cellcolor[HTML]{FFC3C3}55.0 & \cellcolor[HTML]{FFC3C3}54.0 & \cellcolor[HTML]{FFC3C3}{\ul 87.8} & \multicolumn{1}{c|}{\cellcolor[HTML]{FFC3C3}83.0} & \cellcolor[HTML]{FFC3C3}\textbf{94.5} \\ \hline
 & UCE & \multicolumn{1}{c|}{1.0} & 0.0 & 1.0 & {\ul 12.0} & \multicolumn{1}{c|}{24.0} & \textbf{47.0} \\
 & ESD & \multicolumn{1}{c|}{0.0} & 0.0* & 0.0 & {\ul 5.0} & \multicolumn{1}{c|}{2.0} & \textbf{13.0} \\
 & FMN & \multicolumn{1}{c|}{69.0} & 70.0 & 16.0 & \textbf{92.0*} & \multicolumn{1}{c|}{46.0} & {\ul 90.0} \\
 & CA & \multicolumn{1}{c|}{1.0} & 2.0 & 0.0 & 2.0 & \multicolumn{1}{c|}{\textbf{58.0*}} & {\ul 21.0} \\
\multirow{-5}{*}{\begin{tabular}[c]{@{}c@{}}English\\ Springer\\ (Object)\end{tabular}} & \cellcolor[HTML]{FFC3C3}Average & \multicolumn{1}{c|}{\cellcolor[HTML]{FFC3C3}17.8} & \cellcolor[HTML]{FFC3C3}18.0 & \cellcolor[HTML]{FFC3C3}4.3 & \cellcolor[HTML]{FFC3C3}27.8 & \multicolumn{1}{c|}{\cellcolor[HTML]{FFC3C3}{\ul 32.5}} & \cellcolor[HTML]{FFC3C3}\textbf{42.8} \\ \hline
 & UCE & \multicolumn{1}{c|}{0.0} & 0.0 & 1.0 & 1.0 & \multicolumn{1}{c|}{{\ul 26.0}} & \textbf{38.0} \\
 & ESD & \multicolumn{1}{c|}{0.0} & 0.0* & 0.0 & {\ul 9.0} & \multicolumn{1}{c|}{3.0} & \textbf{34.0} \\
 & FMN & \multicolumn{1}{c|}{0.0} & 0.0 & 1.0 & {\ul 51.0*} & \multicolumn{1}{c|}{1.0} & \textbf{54.0} \\
 & CA & \multicolumn{1}{c|}{0.0} & 0.0 & 3.0 & \textbf{60.0} & \multicolumn{1}{c|}{44.0*} & {\ul 55.0} \\
\multirow{-5}{*}{\begin{tabular}[c]{@{}c@{}}Van Gogh\\ (Artist Style)\end{tabular}} & \cellcolor[HTML]{FFC3C3}Average & \multicolumn{1}{c|}{\cellcolor[HTML]{FFC3C3}0.0} & \cellcolor[HTML]{FFC3C3}0.0 & \cellcolor[HTML]{FFC3C3}1.3 & \cellcolor[HTML]{FFC3C3}{\ul 30.3} & \multicolumn{1}{c|}{\cellcolor[HTML]{FFC3C3}18.5} & \cellcolor[HTML]{FFC3C3}\textbf{45.3} \\ \hline
 & UCE & \multicolumn{1}{c|}{0.0} & 0.7 & 0.0 & 1.5 & \multicolumn{1}{c|}{{\ul 5.2}} & \textbf{41.8} \\
 & ESD & \multicolumn{1}{c|}{10.4} & 13.5* & 31.9 & 34.3 & \multicolumn{1}{c|}{{\ul 67.2}} & \textbf{72.4} \\
 & FMN & \multicolumn{1}{c|}{56.0} & 75.9 & 61.7 & 70.9* & \multicolumn{1}{c|}{{\ul 76.1}} & \textbf{87.3} \\
 & CA & \multicolumn{1}{c|}{2.2} & 3.5 & 51.1 & 19.4 & \multicolumn{1}{c|}{\textbf{64.9*}} & {\ul 58.2} \\
\multirow{-5}{*}{\begin{tabular}[c]{@{}c@{}}Nudity\\ (NSFW)\end{tabular}} & \cellcolor[HTML]{FFC3C3}Average & \multicolumn{1}{c|}{\cellcolor[HTML]{FFC3C3}17.2} & \cellcolor[HTML]{FFC3C3}23.4 & \cellcolor[HTML]{FFC3C3}36.2 & \cellcolor[HTML]{FFC3C3}31.5 & \multicolumn{1}{c|}{\cellcolor[HTML]{FFC3C3}{\ul 53.4}} & \cellcolor[HTML]{FFC3C3}\textbf{64.9} \\ \hline
 & UCE & \multicolumn{1}{c|}{0.0} & 0.0 & 0.0 & 0.0 & \multicolumn{1}{c|}{{\ul 10.0}} & \textbf{39.0} \\
 & ESD & \multicolumn{1}{c|}{0.0} & 0.0* & 0.0 & {\ul 5.0} & \multicolumn{1}{c|}{0.0} & \textbf{32.0} \\
 & FMN & \multicolumn{1}{c|}{0.0} & 0.0 & 0.0 & {\ul 56.0*} & \multicolumn{1}{c|}{0.0} & \textbf{81.0} \\
 & CA & \multicolumn{1}{c|}{0.0} & 0.0 & 0.0 & {\ul 47.0} & \multicolumn{1}{c|}{41.0*} & \textbf{70.0} \\
\multirow{-5}{*}{\begin{tabular}[c]{@{}c@{}}Barack\\ Obama\\ (ID)\end{tabular}} & \cellcolor[HTML]{FFC3C3}Average & \multicolumn{1}{c|}{\cellcolor[HTML]{FFC3C3}0.0} & \cellcolor[HTML]{FFC3C3}0.0 & \cellcolor[HTML]{FFC3C3}0.0 & \cellcolor[HTML]{FFC3C3}{\ul 27.0} & \multicolumn{1}{c|}{\cellcolor[HTML]{FFC3C3}12.8} & \cellcolor[HTML]{FFC3C3}\textbf{55.5} \\ \Xhline{1.0pt}
\end{tabular}
\vspace{-0.4cm}
\end{table}

\begin{figure*}
\centering
\includegraphics[width=0.93\textwidth]{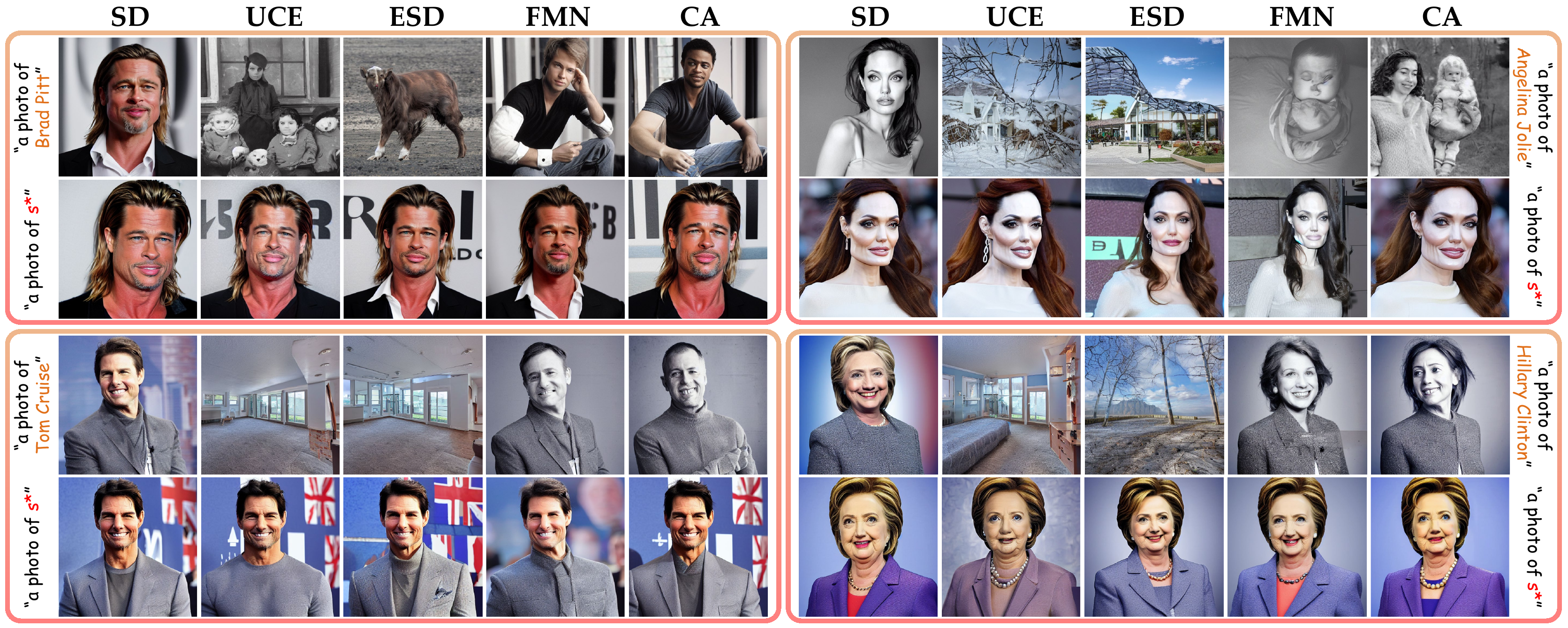}
\caption{More results for restoring identities on unlearned models using our method.}
\label{figure_ours_id}
\vspace{-0.2cm}
\end{figure*}

\begin{figure*}
\centering
\includegraphics[width=0.93\textwidth]{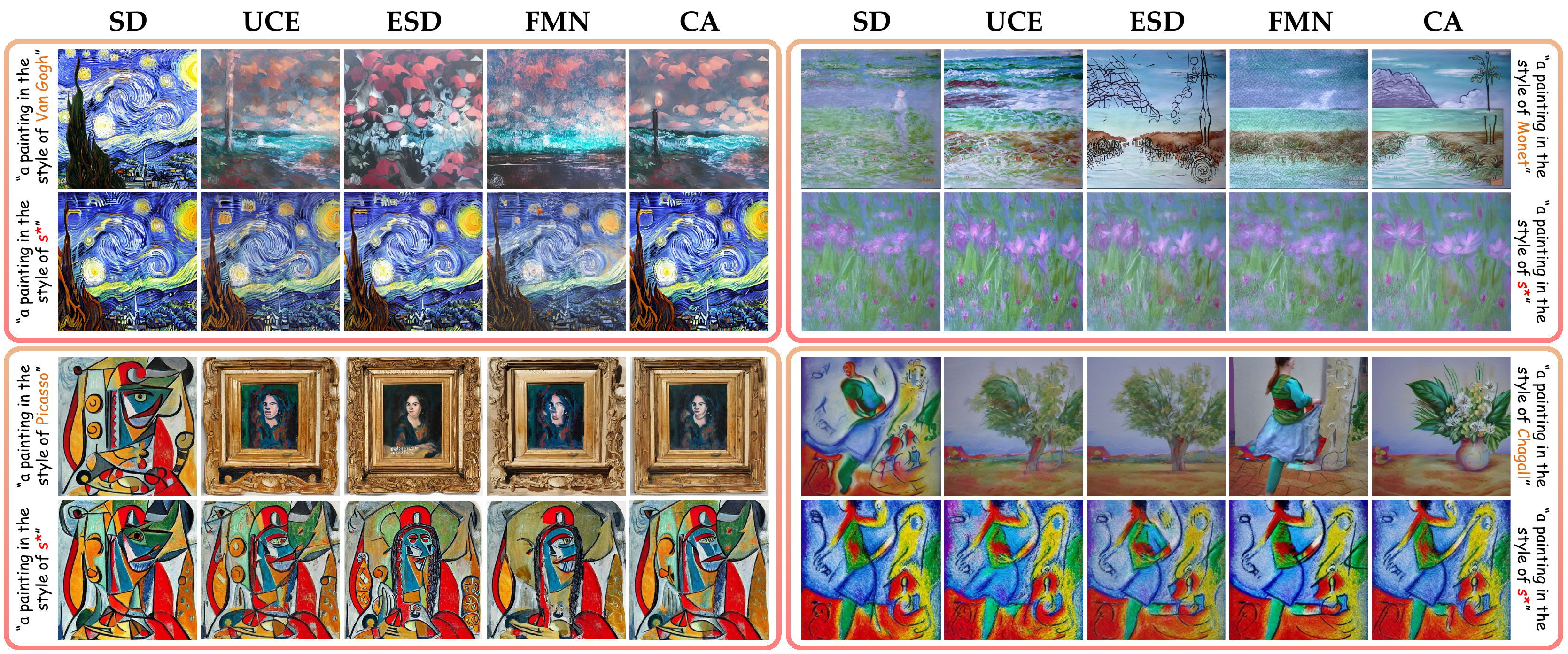}
\caption{More results for restoring artist styles on unlearned models using our method.}
\label{figure_ours_style}
\vspace{-0.2cm}
\end{figure*}

\section{Experiments}
\subsection{Experimental Setup}

\textbf{Implementation Details.} The text-to-image model used in this paper is Stable Diffusion (SD) v1.4, chosen due to its widespread adoption in most concept erasure methods \cite{esd,ca,uce}. VIT-L/14 \cite{clip} serves as the text encoder. We probe the robustness of four representative erasure methods, namely UCE \cite{uce}, ESD \cite{esd}, FMN \cite{fmn} and CA \cite{ca}. These methods were re-implemented using the official code, with detailed hyperparameters provided in the Appendix.

\noindent\textbf{Baseline Methods.} We compare the restoration performance of the proposed method with three baseline approaches: UnlearnDiff (UD) \cite{unlearndiff}, Ring-A-Bell (RAB) \cite{ringabell}, and the Concept Inversion (CI) techniques used by Pham et al. \cite{circumventing}. UD and CI are white-box attack methods. For UD, we employ the ESD erased model for optimization, and for CI, we use the FMN and CA erased model respectively.

\noindent\textbf{Metrics.} We employ specific classifiers to detect the target concept within the generated images. A higher classification accuracy indicates a greater likelihood of the target concept being present in the generated images. Concretely, to assess the restoration of object concepts, we use YOLOv3 \cite{yolov3} to detect broad concepts (e.g., ``Dog'') and ResNet-18 \cite{resnet} trained on ImageNet \cite{imagenet} to identify narrow concepts (e.g., ``English Springer''). For artist style probe, we utilize the classifier provided by \cite{unlearndiff}. For the probe of NSFW (i.e., nudity) content, we use Nudenet \cite{nudenet} to detect nudity. In assessing the restoration of celebrity identity (ID) concepts, we use the GIPHY Celebrity Detector \cite{giphy} to classify the person in the generated images. The detailed configurations of these classifiers (e.g., detection score threshold for Nudenet) are provided in the Appendix.

\begin{table*}
\footnotesize
\caption{Restoration Performance of More Artist Styles and Identities Using the Proposed Method.}
\label{table_example}
\begin{tabular}{cc|cccccccccc}
\Xhline{1.0pt}
\multicolumn{2}{c|}{\textbf{Target Concepts $\downarrow$}} & \multicolumn{2}{c}{SD} & \multicolumn{2}{c}{UCE} & \multicolumn{2}{c}{ESD} & \multicolumn{2}{c}{FMN} & \multicolumn{2}{c}{CA} \\ \hline
\multicolumn{1}{c|}{Type} & Example & \begin{tabular}[c]{@{}c@{}}w/o\\ attack\end{tabular} & \begin{tabular}[c]{@{}c@{}}w/\\ attack\end{tabular} & \begin{tabular}[c]{@{}c@{}}w/o\\ attack\end{tabular} & \begin{tabular}[c]{@{}c@{}}w/\\ attack\end{tabular} & \begin{tabular}[c]{@{}c@{}}w/o\\ attack\end{tabular} & \begin{tabular}[c]{@{}c@{}}w/\\ attack\end{tabular} & \begin{tabular}[c]{@{}c@{}}w/o\\ attack\end{tabular} & \begin{tabular}[c]{@{}c@{}}w/\\ attack\end{tabular} & \begin{tabular}[c]{@{}c@{}}w/o\\ attack\end{tabular} & \begin{tabular}[c]{@{}c@{}}w/\\ attack\end{tabular} \\ \Xhline{1.0pt}
\multicolumn{1}{c|}{} & Monet & 100.0 & 50.0 & 2.5 & 29.0 & 0.0 & 9.0 & 0.5 & 14.0 & 0.0 & 11.0 \\
\multicolumn{1}{c|}{} & Pablo Picasso & 30.0 & 32.5 & 0.0 & 16.0 & 0.0 & 17.5 & 0.0 & 9.5 & 0.0 & 40.5 \\
\multicolumn{1}{c|}{} & Marc Chagall & 99.5 & 89.0 & 0.5 & 49.0 & 0.0 & 61.5 & 0.5 & 76.5 & 1.0 & 86.0 \\
\multicolumn{1}{c|}{\multirow{-4}{*}{\begin{tabular}[c]{@{}c@{}}Artist\\ Style\end{tabular}}} & \cellcolor[HTML]{FFC3C3}Average & \cellcolor[HTML]{FFC3C3}82.4 & \cellcolor[HTML]{FFC3C3}60.0 & \cellcolor[HTML]{FFC3C3}0.8 & \cellcolor[HTML]{FFC3C3}32.5 & \cellcolor[HTML]{FFC3C3}0.0 & \cellcolor[HTML]{FFC3C3}31.0 & \cellcolor[HTML]{FFC3C3}0.3 & \cellcolor[HTML]{FFC3C3}38.9 & \cellcolor[HTML]{FFC3C3}0.4 & \cellcolor[HTML]{FFC3C3}48.1 \\ \hline
\multicolumn{1}{c|}{} & Emma Watson & 99.0 & 89.5 & 0.0 & 56.5 & 0.0 & 69.0 & 3.0 & 70.5 & 0.0 & 69.5 \\
\multicolumn{1}{c|}{} & Brad Pitt & 100.0 & 98.0 & 0.0 & 90.0 & 0.0 & 97.0 & 0.0 & 81.5 & 0.5 & 96.5 \\
\multicolumn{1}{c|}{} & Angelina Jolie & 100.0 & 99.5 & 0.0 & 93.5 & 0.0 & 85.0 & 0.0 & 70.5 & 1.0 & 99.0 \\
\multicolumn{1}{c|}{} & Tom Cruise & 100.0 & 84.5 & 0.0 & 65.5 & 0.0 & 28.5 & 0.5 & 23.0 & 0.0 & 70.5 \\
\multicolumn{1}{c|}{} & Hillary Clinton & 100.0 & 82.5 & 0.0 & 49.0 & 0.0 & 37.0 & 3.0 & 73.0 & 1.0 & 85.0 \\
\multicolumn{1}{c|}{\multirow{-6}{*}{ID}} & \cellcolor[HTML]{FFC3C3}Average & \cellcolor[HTML]{FFC3C3}99.8 & \cellcolor[HTML]{FFC3C3}85.8 & \cellcolor[HTML]{FFC3C3}0.0 & \cellcolor[HTML]{FFC3C3}65.3 & \cellcolor[HTML]{FFC3C3}0.0 & \cellcolor[HTML]{FFC3C3}57.6 & \cellcolor[HTML]{FFC3C3}1.3 & \cellcolor[HTML]{FFC3C3}66.0 & \cellcolor[HTML]{FFC3C3}0.4 & \cellcolor[HTML]{FFC3C3}81.3 \\ \Xhline{1.0pt}
\end{tabular}

\end{table*}

\subsection{Comparisons with Baseline Methods}

\noindent\textbf{Object.} We first probe the restoration of objects concepts, encompassing both broad objects (e.g., ``Dog'') and narrow ones (e.g., ``English Springer''). The restoration performance of the baseline methods and ours is presented in Table \ref{table_cmp} and Figure~\ref{figure_object}. For broad objects, prompt-level attacks (i.e., UD \cite{unlearndiff} and RAB \cite{ringabell}) are effective for the restoration, while embedding-level attacks yield stronger results. But for narrow object, prompt-level attacks can not effectively restore the erased concept. This indicates that narrower concepts are easier to erase and harder to restore. Although the white-box embedding-level attack \cite{circumventing} can restore narrow object for the known erasure method, its transferability is limited. Conversely, our black-box method exhibits superior transferability across various erasure methods. Additional restoration results of more objects are provided in the Appendix.

\noindent\textbf{Artist Style.}
We then probe the restoration of artist style. As shown in Table \ref{table_cmp} and Figure~\ref{figure_nsfw+style}, prompt-level attacks \cite{unlearndiff,ringabell} are ineffective for the artist style restoration, and the white-box embedding-level attack exhibits poor transferability to other erasure methods. But ours achieves better performance. Our method's restoration performance of more artist styles can be found in Table \ref{table_example} and Figure ~\ref{figure_ours_style}.

\begin{figure}
\centering
\includegraphics[width=0.44\textwidth]{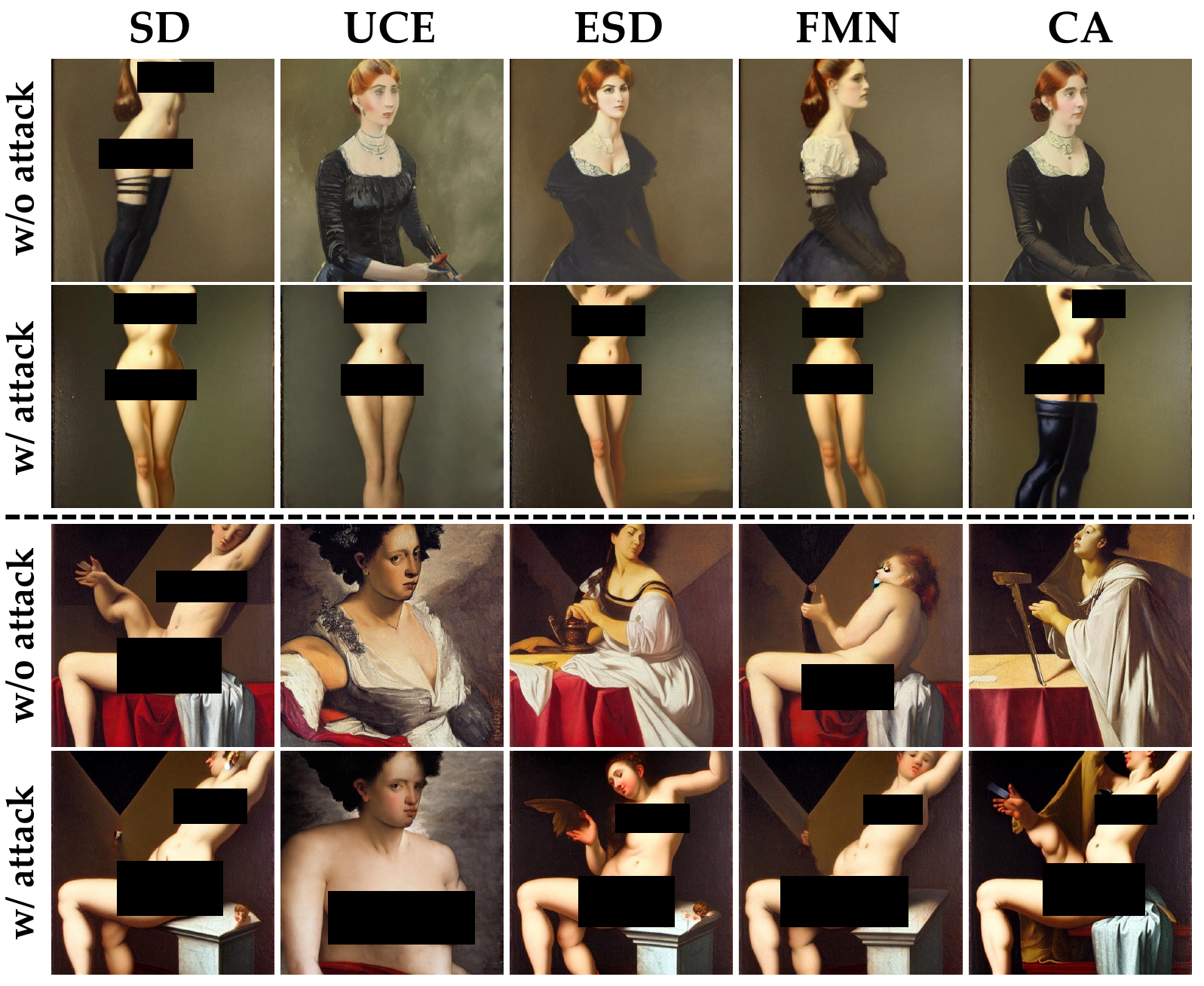}
\caption{More results for restoring NSFW content.}
\label{figure_ours_nsfw}
\vspace{-0.4cm}
\end{figure}

\noindent\textbf{NSFW Content.}
As depicted in Table \ref{table_cmp} and Figure \ref{figure_nsfw+style}, regarding the restoration of nudity content, prompt-level attacks are effective for some erasure methods but may fail when applied to specific erasure methods (e.g., UCE). Similarly, the white-box embedding-level attack also falls short in restoring nudity content across all erasure methods. On the contrary, our method demonstrates effective restoration of nudity content across all four erasure methods, with additional qualitative results shown in Figure ~\ref{figure_ours_nsfw}.

\noindent\textbf{Identity.} Lastly, we probe the restoration of celebrity identity (ID) concepts, which is the most challenging type. From the results shown in Table \ref{table_cmp} and Figure \ref{figure_id}, we can find that the prompt-level attacks fail to restore the target ID. Additionally, the white-box embedding-level attack \cite{circumventing} is primarily effective against known erasure methods and struggles to transfer to others. In contrast, our approach can still restore the target ID under the black-box setting, demonstrating its strong transferability. We also use the proposed method to restore diverse ID concepts, and the quantitative and qualitative results are shown in Table \ref{table_example} and Figure \ref{figure_ours_id}, respectively.

\begin{figure}
\centering
\includegraphics[width=0.43\textwidth]{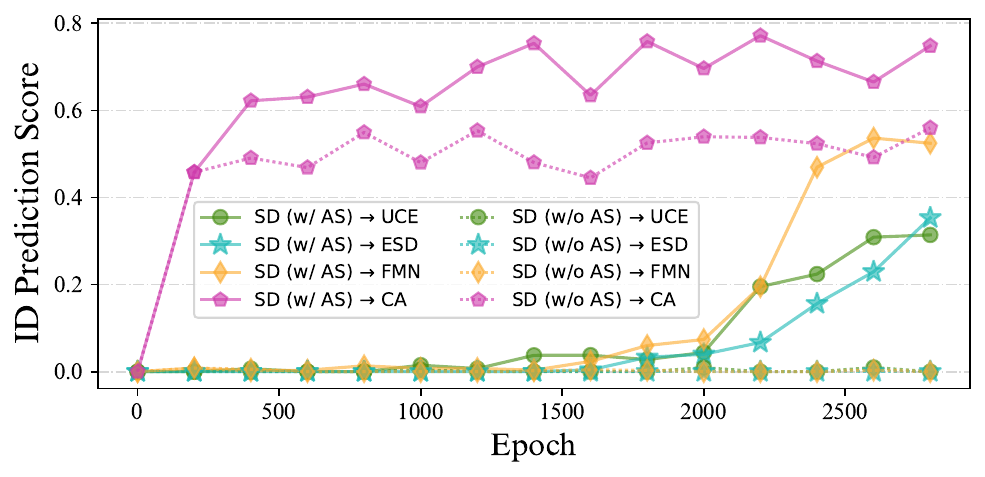}
\vspace{-0.1cm}
\caption{Ablation study of Adversarial Search (AS) strategy.}
\label{figure_ablation}
\vspace{-0.4cm}
\end{figure}

\subsection{Ablation Study} 
 
Utilizing the original Stable Diffusion (SD) model, we search for the adversarial embedding for ID restoration with and without Adversarial Search (AS) strategy, respectively. The embeddings obtained during the optimization process are then fed into various unlearned models to generate images. We record the average ID prediction score every 200 epochs to track restoration performance throughout the optimization process. The results are depicted in Figure~\ref{figure_ablation}. It is evident that, for most erasure methods, embeddings obtained without AS struggle to restore the target ID. Conversely, with the assistance of AS, the ID prediction score gradually increases. Additionally, embeddings obtained without AS manage to restore the target ID for the model erased by CA, consistent with the findings in Figure~\ref{figure_tsne}, where embeddings obtained from SD without AS exhibit significant overlap with those obtained from CA.

\section{Conclusions}
In this paper, we propose an adversarial search strategy to find the transferable embedding for probing erasure robustness under a black-box setting. This strategy alternately erases and searches for embeddings, enabling it to find embeddings that can restore the target concept for various unlearning methods. Extensive experiments demonstrate the transferability of the acquired adversarial embedding across several state-of-the-art unlearning methods and its effectiveness across different levels of concepts, including objects, artist styles, NSFW content, and the most challenging identity.

\noindent\textbf{Ethical Statement.} Our work is of the utmost importance for content security. Investigating concept restoration enables us to uncover vulnerabilities in existing concept erasure methods. We are committed to further expanding this work to develop more robust concept erasure techniques.

\bibliographystyle{ACM-Reference-Format}
\bibliography{sample-base}

\appendix

\section{APPENDIX}
\subsection{Implementation Details}
\noindent \textbf{Hyperparameters of Erasure Methods.} This paper investigates four erasure methods. Unified Concept Editing (UCE) \cite{uce} directs the target concepts towards the unconditional concept (i.e., `` ''). For Erased Stable Diffusion (ESD) \cite{esd}, we fine-tune cross-attention parameters (ESD-x) for object, artist style, and identity (ID) erasure over 1000 iterations with a learning rate of 1e-5. For NSFW content erasure, we fine-tune unconditional layers (ESD-u) for 1000 iterations with a learning rate of 1e-5. The negative guidance is set to 1.0 for all concepts. For Forget-Me-Not (FMN) \cite{fmn}, object erasure involves 50 training steps with a learning rate of 2e-6, identity erasure comprises 35 training steps with the same learning rate, artist style erasure utilizes 35 training steps with a learning rate of 1e-5, and nudity erasure employs a checkpoint provided by \cite{unlearndiff}. For Concept Ablation (CA) \cite{ca}, object, nudity, and identity erasure involve 500 training steps with a learning rate of 2e-5, while artist style erasure comprises 200 training steps with a learning rate of 1e-5.

\noindent \textbf{Hyperparameters of Restoration Methods.} 
We investigate four restoration methods, consisting of three baseline methods along with the proposed one. For UnlearnDiff (UD) \cite{unlearndiff}, adversarial tokens are placed at the sentence prefix. For object and identity concepts, 3 adversarial tokens are used; for NSFW content and artist style, 5 tokens are employed. Learning rate is 0.01 and weight decay is 0.1. For Ring-A-Bell (RAB) \cite{ringabell}, 30 prompt pairs are utilized to obtain the concept vector. The length of prompts is 16 and the strength coefficient is 3. For Concept Inversion (CI) \cite{circumventing}, the learning rate for updating adversarial embeddings is 0.1, and weight decay is 0.1. For the proposed method, when updating the adversarial embedding, the learning rate and weight decay are both set to 0.1. When updating the model parameters, the learning rate is set to 1e-5.

\noindent \textbf{Evaluation Details.}
Broad object (e.g., car) restoration is evaluated using YOLOv3 \cite{yolov3} with a score threshold of 0.5 and a Non-Maximum Suppression (NMS) threshold of 0.4. Narrow object (e.g., jeep) restoration is evaluated using the pre-trained ResNet-18 \cite{resnet} provided by torchvision library. Nudity restoration is evaluated with Nudenet \cite{nudenet} by detecting five types: BUTTOCKS\_EXPOSED, FEMALE\_BREAST\_EXPOSED, FEMALE\_GENITALIA\_EXPOSED, ANUS\_EXPOSED and MALE\_GENITALIA\_EXPOSED, with a score threshold of 0.6. Restoration of artist styles is evaluated using the model provided by \cite{unlearndiff}, with performance measured by Top-1 accuracy.

\subsection{Additional Results}

\begin{figure*}
\centering
\includegraphics[width=0.99\textwidth]{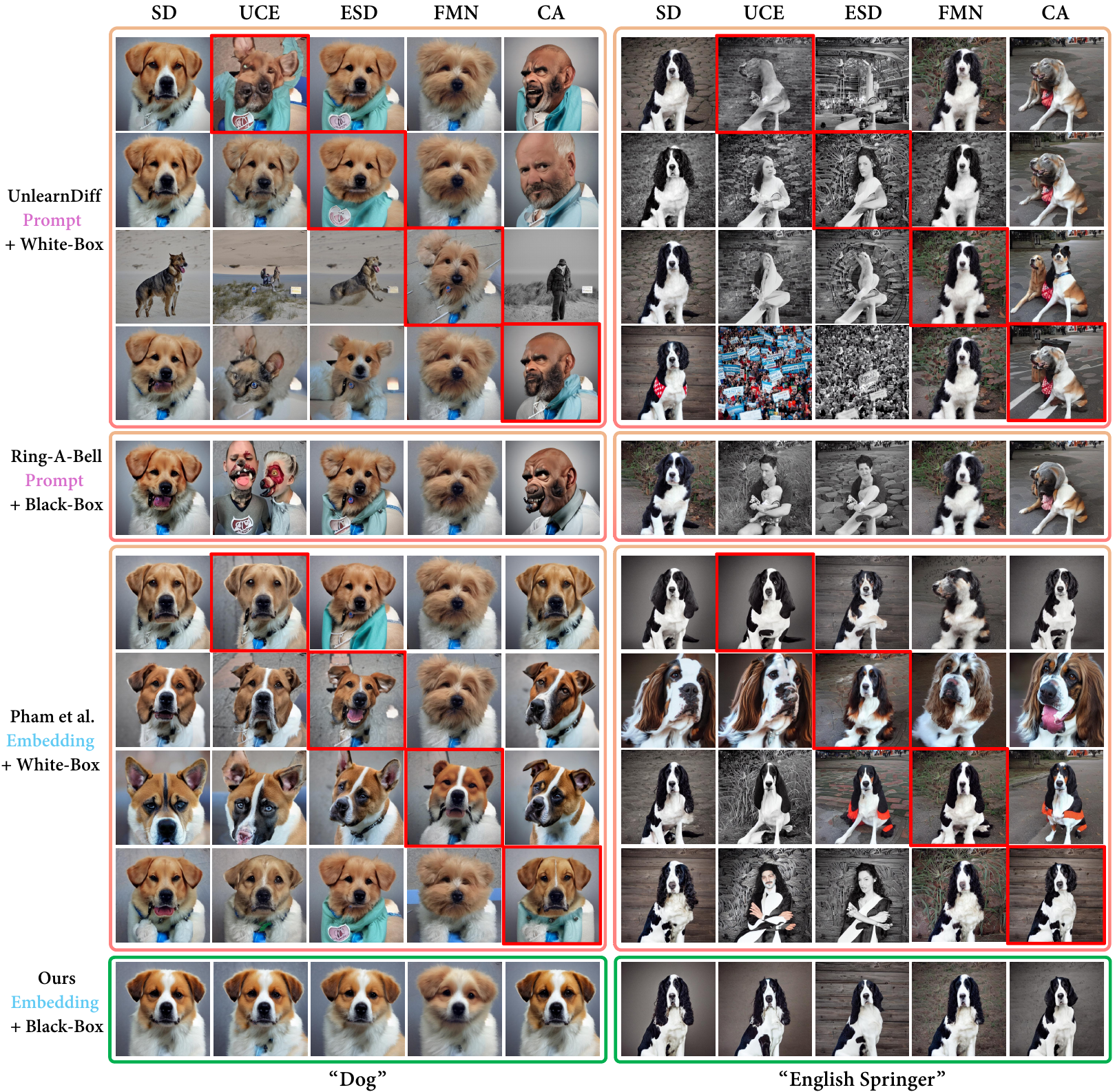}
\caption{Additional comparison results for objects restoration using different concept restoration methods (the \textcolor{red}{red} border represents that the unlearned model is accessible to the attacker).}
\label{figure_more_objects_dog}
\end{figure*}

\begin{figure*}
\centering
\includegraphics[width=0.99\textwidth]{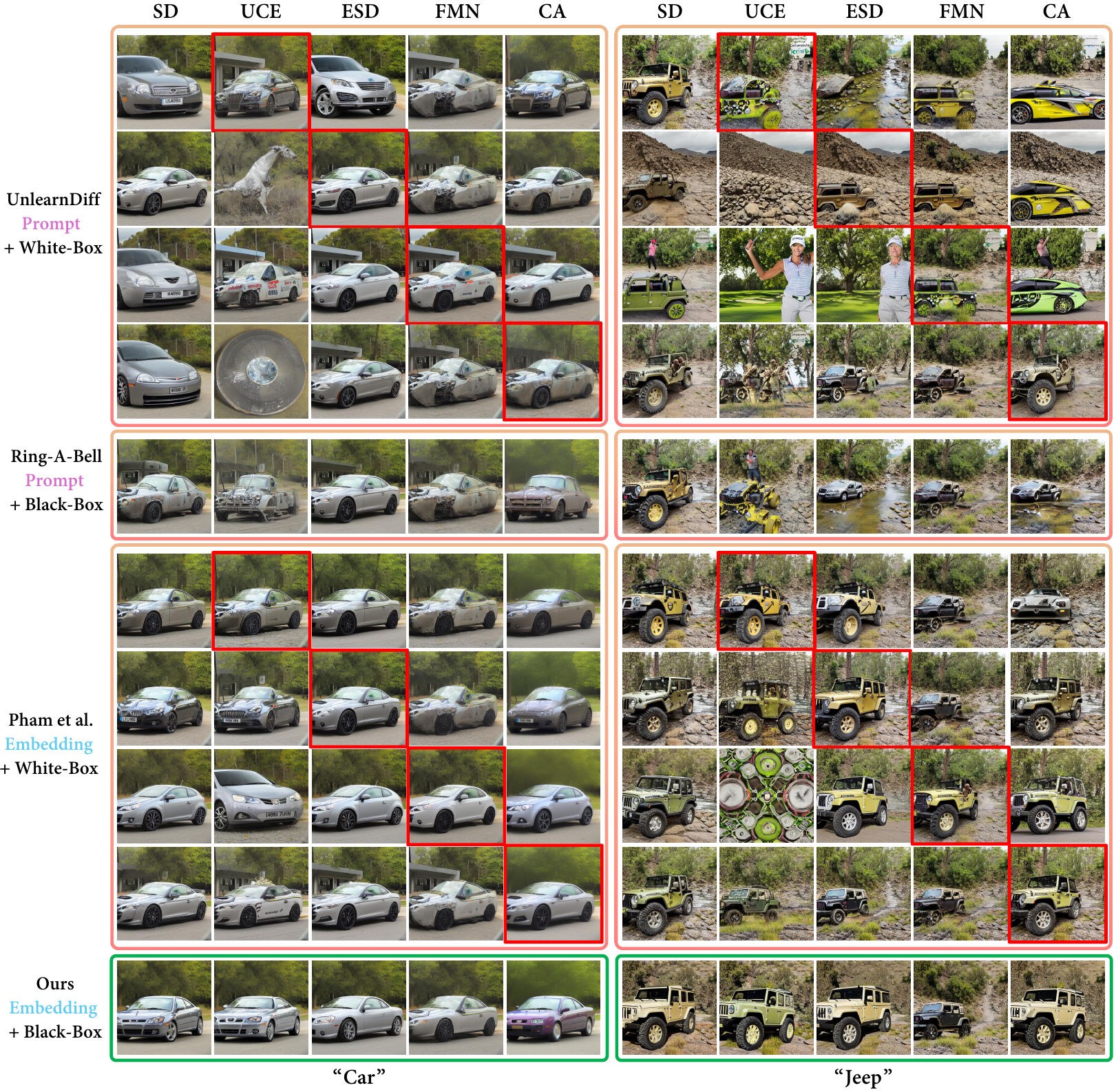}
\caption{Additional comparison results for objects restoration using different concept restoration methods (the \textcolor{red}{red} border represents that the unlearned model is accessible to the attacker).}
\label{figure_more_objects_car}
\end{figure*}

\begin{figure*}
\centering
\includegraphics[width=0.99\textwidth]{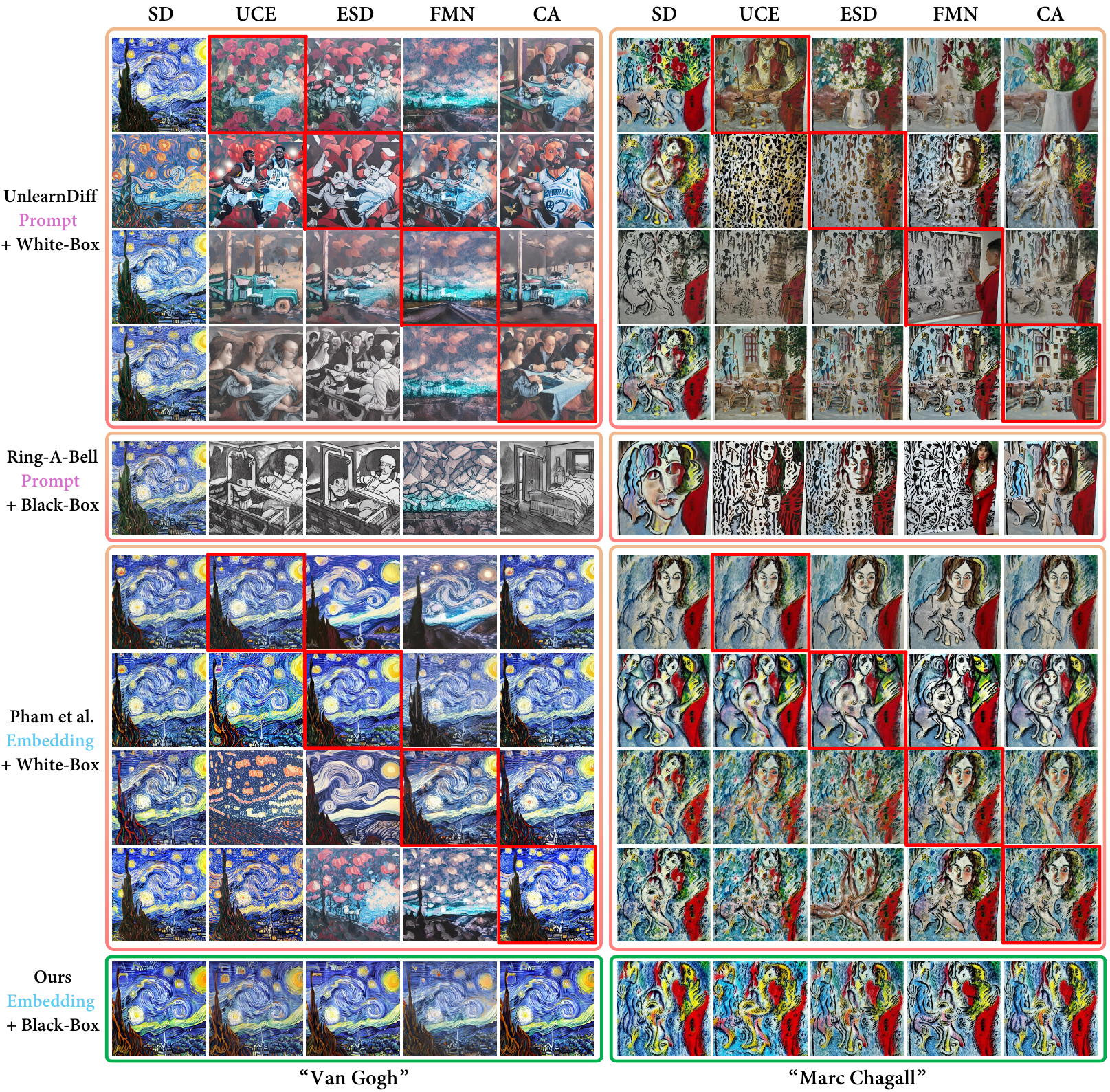}
\caption{Additional comparison results for artist styles restoration using different concept restoration methods (the \textcolor{red}{red} border represents that the unlearned model is accessible to the attacker).}
\label{figure_more_styles}
\end{figure*}

\begin{figure*}
\centering
\includegraphics[width=0.99\textwidth]{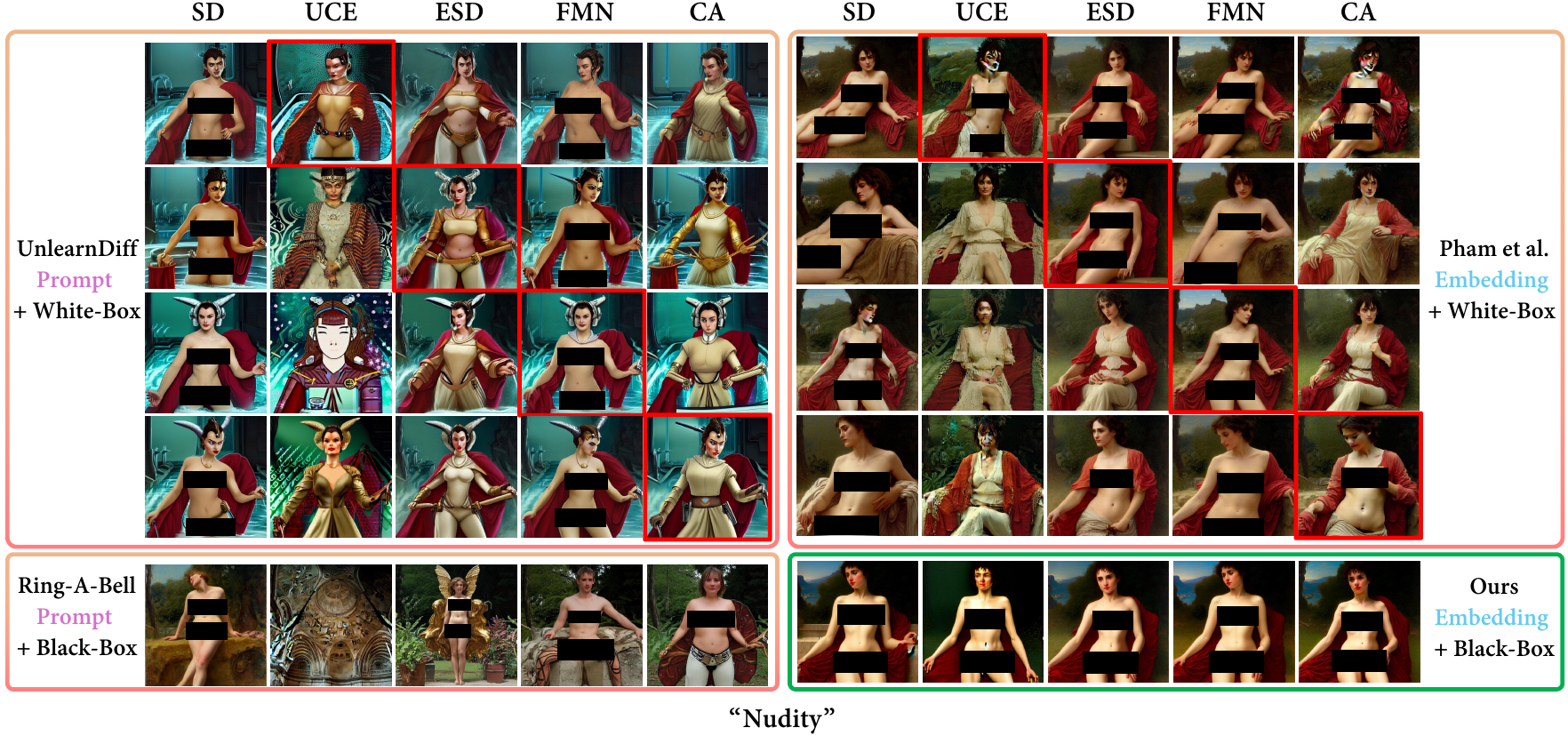}
\caption{Additional comparison results for NSFW content restoration using different concept restoration methods (the \textcolor{red}{red} border represents that the unlearned model is accessible to the attacker).}
\label{figure_more_nudity}
\end{figure*}

\begin{figure*}
\centering
\includegraphics[width=0.99\textwidth]{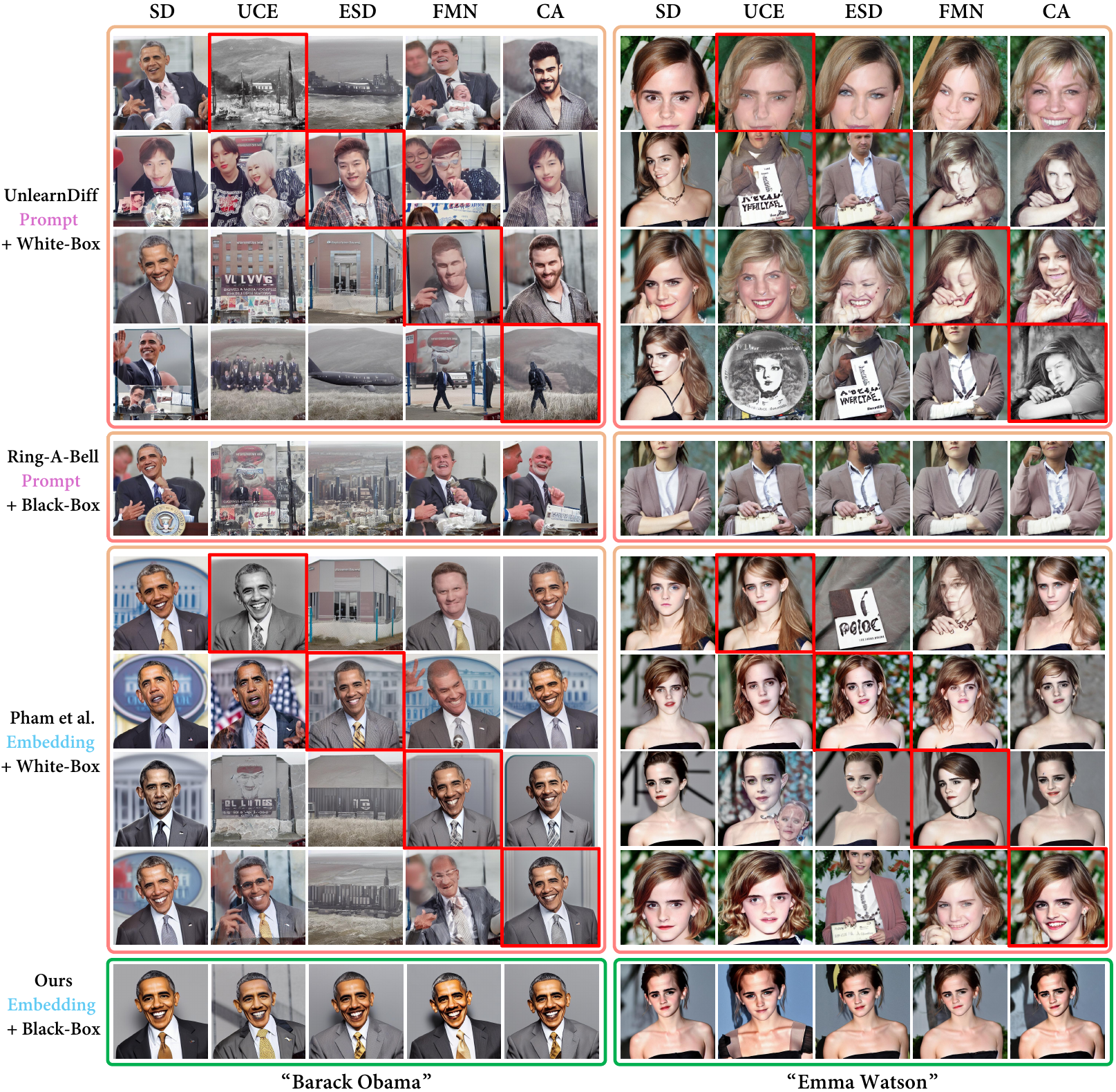}
\caption{Additional comparison results for celebrity identities restoration using different concept restoration methods (the \textcolor{red}{red} border represents that the unlearned model is accessible to the attacker).}
\label{figure_more_ids}
\end{figure*}

In the manuscript, for the white-box restoration method UnlearnDiff (UD) \cite{unlearndiff}, we utilize ESD \cite{esd} as the known erasure method. Here, we present additional results for UD when UCE \cite{uce}, FMN \cite{fmn}, and CA \cite{ca} are the known erasure methods (i.e., UD (UCE), UD (FMN), and UD (CA)). Similarly, for the other white-box restoration method Concept Inversion (CI) \cite{circumventing}, we introduce results when UCE \cite{uce} and ESD \cite{esd} are the known erasure methods (i.e., CI (UCE) and CI (ESD)). In addition to introducing more variants of baseline methods, we also include more concept examples. Specifically, for objects, we introduce comparison results for ``Car'' and ``Jeep''. Regarding artist style, we include comparison results for ``Marc Chagall'', and for celebrity identity, we add the comparison results for ``Emma Watson''. The qualitative results for objects restoration are depicted in Figure \ref{figure_more_objects_dog} and Figure \ref{figure_more_objects_car}, while those for artist styles restoration are shown in Figure \ref{figure_more_styles}. NSFW restoration results are presented in Figure \ref{figure_more_nudity}, and celebrity identity restoration results are shown in Figure \ref{figure_more_ids}. The quantitative results of all these concepts are presented in Table \ref{table_more_cmp}. The proposed method achieves superior restoration performance for each concept across various erasure methods, as evidenced by higher average accuracy, highlighting its enhanced transferability.

\begin{table*}[]
\caption{More Comparisons of Different Attack Methods on Diverse Concepts. The best results in \textbf{bold}, the second best \underline{underlined}. The asterisk (*) denotes a white-box attack.}
\label{table_more_cmp}
\scalebox{0.95}{
\begin{tabular}{c|c|ccccccccccc}
\Xhline{1.0pt}
 &  & \multicolumn{11}{c}{\textbf{Attack Methods}} \\ \cline{3-13} 
\multirow{-2}{*}{\textbf{\begin{tabular}[c]{@{}c@{}}Target\\ Concepts\end{tabular}}} & \multirow{-2}{*}{\textbf{\begin{tabular}[c]{@{}c@{}}Erasure\\ Methods\end{tabular}}} & \multicolumn{1}{c|}{\begin{tabular}[c]{@{}c@{}}w/o \\ Attack\end{tabular}} & \begin{tabular}[c]{@{}c@{}}UD\\ (UCE)\end{tabular} & \begin{tabular}[c]{@{}c@{}}UD\\ (ESD)\end{tabular} & \begin{tabular}[c]{@{}c@{}}UD\\ (FMN)\end{tabular} & \begin{tabular}[c]{@{}c@{}}UD\\ (CA)\end{tabular} & RAB & \begin{tabular}[c]{@{}c@{}}CI\\ (UCE)\end{tabular} & \begin{tabular}[c]{@{}c@{}}CI\\ (ESD)\end{tabular} & \begin{tabular}[c]{@{}c@{}}CI\\ (FMN)\end{tabular} & \multicolumn{1}{c|}{\begin{tabular}[c]{@{}c@{}}CI\\ (CA)\end{tabular}} & Ours \\ \Xhline{1.0pt}
 & UCE & \multicolumn{1}{c|}{0.0} & 29.0* & 37.0 & 1.0 & 16.0 & 29.0 & 82.0* & 78.0 & {\ul 86.0} & \multicolumn{1}{c|}{82.0} & \textbf{96.0} \\
 & ESD & \multicolumn{1}{c|}{16.0} & 66.0 & 57.0* & 25.0 & 43.0 & 62.0 & 80.0 & {\ul 94.0*} & 93.0 & \multicolumn{1}{c|}{78.0} & \textbf{98.0} \\
 & FMN & \multicolumn{1}{c|}{65.0} & 72.0 & 64.0 & 70.0* & 71.0 & 66.0 & 70.0 & 70.0 & {\ul 82.0*} & \multicolumn{1}{c|}{74.0} & \textbf{84.0} \\
 & CA & \multicolumn{1}{c|}{12.0} & 44.0 & 62.0 & 14.0 & 52.0* & 59.0 & 91.0 & 91.0 & 90.0 & \multicolumn{1}{c|}{{\ul 98.0*}} & \textbf{100.0} \\
\multirow{-5}{*}{\begin{tabular}[c]{@{}c@{}}Dog\\ (Object)\end{tabular}} & \cellcolor[HTML]{FFC3C3}Average & \multicolumn{1}{c|}{\cellcolor[HTML]{FFC3C3}23.3} & \cellcolor[HTML]{FFC3C3}52.8 & \cellcolor[HTML]{FFC3C3}55.0 & \cellcolor[HTML]{FFC3C3}27.5 & \cellcolor[HTML]{FFC3C3}45.5 & \cellcolor[HTML]{FFC3C3}54.0 & \cellcolor[HTML]{FFC3C3}80.8 & \cellcolor[HTML]{FFC3C3}83.3 & \cellcolor[HTML]{FFC3C3}{\ul 87.8} & \multicolumn{1}{c|}{\cellcolor[HTML]{FFC3C3}83.0} & \cellcolor[HTML]{FFC3C3}\textbf{94.5} \\ \hline
 & UCE & \multicolumn{1}{c|}{1.0} & 1.0* & 0.0 & 0.0 & 0.0 & 1.0 & \textbf{51.0*} & 41.0 & 12.0 & \multicolumn{1}{c|}{24.0} & {\ul 47.0} \\
 & ESD & \multicolumn{1}{c|}{0.0} & 0.0 & 0.0* & 0.0 & 0.0 & 0.0 & 8.0 & \textbf{27.0*} & 5.0 & \multicolumn{1}{c|}{2.0} & {\ul 13.0} \\
 & FMN & \multicolumn{1}{c|}{69.0} & 60.0 & 70.0 & 70.0* & 50.0 & 16.0 & 16.0 & 50.0 & \textbf{92.0*} & \multicolumn{1}{c|}{46.0} & {\ul 90.0} \\
 & CA & \multicolumn{1}{c|}{1.0} & 2.0 & 2.0 & 2.0 & 1.0* & 0.0 & {\ul 28.0} & 16.0 & 2.0 & \multicolumn{1}{c|}{\textbf{58.0*}} & 21.0 \\
\multirow{-5}{*}{\begin{tabular}[c]{@{}c@{}}English\\ Springer\\ (Object)\end{tabular}} & \cellcolor[HTML]{FFC3C3}Average & \multicolumn{1}{c|}{\cellcolor[HTML]{FFC3C3}17.8} & \cellcolor[HTML]{FFC3C3}15.8 & \cellcolor[HTML]{FFC3C3}18.0 & \cellcolor[HTML]{FFC3C3}18.0 & \cellcolor[HTML]{FFC3C3}12.8 & \cellcolor[HTML]{FFC3C3}4.3 & \cellcolor[HTML]{FFC3C3}25.8 & \cellcolor[HTML]{FFC3C3}{\ul 33.5} & \cellcolor[HTML]{FFC3C3}27.8 & \multicolumn{1}{c|}{\cellcolor[HTML]{FFC3C3}32.5} & \cellcolor[HTML]{FFC3C3}\textbf{42.8} \\ \hline
 & UCE & \multicolumn{1}{c|}{16.0} & 19.0* & 7.0 & 22.0 & 6.0 & 21.0 & \textbf{99.0*} & 55.0 & 89.0 & \multicolumn{1}{c|}{57.0} & {\ul 98.0} \\
 & ESD & \multicolumn{1}{c|}{21.0} & 64.0 & 46.0* & 62.0 & 52.0 & 51.0 & 91.0 & 94.0* & {\ul 98.0} & \multicolumn{1}{c|}{69.0} & \textbf{99.0} \\
 & FMN & \multicolumn{1}{c|}{72.0} & 72.0 & 59.0 & 82.0* & 59.0 & 67.0 & 81.0 & {\ul 92.0} & \textbf{99.0*} & \multicolumn{1}{c|}{81.0} & 91.0 \\
 & CA & \multicolumn{1}{c|}{80.0} & 85.0 & 71.0 & 83.0 & 78.0* & 54.0 & 89.0 & {\ul 99.0} & \textbf{100.0} & \multicolumn{1}{c|}{93.0*} & \textbf{100.0} \\
\multirow{-5}{*}{\begin{tabular}[c]{@{}c@{}}Car\\ (Object)\end{tabular}} & \cellcolor[HTML]{FFC3C3}Average & \multicolumn{1}{c|}{\cellcolor[HTML]{FFC3C3}47.3} & \cellcolor[HTML]{FFC3C3}60.0 & \cellcolor[HTML]{FFC3C3}45.8 & \cellcolor[HTML]{FFC3C3}62.3 & \cellcolor[HTML]{FFC3C3}48.8 & \cellcolor[HTML]{FFC3C3}48.3 & \cellcolor[HTML]{FFC3C3}90.0 & \cellcolor[HTML]{FFC3C3}85.0 & \cellcolor[HTML]{FFC3C3}{\ul 96.5} & \multicolumn{1}{c|}{\cellcolor[HTML]{FFC3C3}75.0} & \cellcolor[HTML]{FFC3C3}\textbf{97.0} \\ \hline
 & UCE & \multicolumn{1}{c|}{4.0} & 11.0* & 24.0 & 0.0 & 2.0 & 30.0 & \textbf{99.0*} & {\ul 81.0} & 0.0 & \multicolumn{1}{c|}{73.0} & \textbf{99.0} \\
 & ESD & \multicolumn{1}{c|}{4.0} & 7.0 & 32.0* & 3.0 & 27.0 & 28.0 & 83.0 & {\ul 98.0*} & 85.0 & \multicolumn{1}{c|}{78.0} & \textbf{99.0} \\
 & FMN & \multicolumn{1}{c|}{39.0} & 39.0 & 51.0 & 16.0* & 55.0 & 62.0 & 78.0 & {\ul 91.0} & \textbf{97.0*} & \multicolumn{1}{c|}{83.0} & \textbf{97.0} \\
 & CA & \multicolumn{1}{c|}{0.0} & 0.0 & 13.0 & 2.0 & 61.0* & 45.0 & 70.0 & {\ul 98.0} & 88.0 & \multicolumn{1}{c|}{\textbf{99.0*}} & \textbf{99.0} \\
\multirow{-5}{*}{\begin{tabular}[c]{@{}c@{}}Jeep\\ (Object)\end{tabular}} & \cellcolor[HTML]{FFC3C3}Average & \multicolumn{1}{c|}{\cellcolor[HTML]{FFC3C3}11.8} & \cellcolor[HTML]{FFC3C3}14.3 & \cellcolor[HTML]{FFC3C3}30.0 & \cellcolor[HTML]{FFC3C3}5.3 & \cellcolor[HTML]{FFC3C3}36.3 & \cellcolor[HTML]{FFC3C3}41.3 & \cellcolor[HTML]{FFC3C3}82.5 & \cellcolor[HTML]{FFC3C3}{\ul 92.0} & \cellcolor[HTML]{FFC3C3}67.5 & \multicolumn{1}{c|}{\cellcolor[HTML]{FFC3C3}83.3} & \cellcolor[HTML]{FFC3C3}\textbf{98.5} \\ \hline
 & UCE & \multicolumn{1}{c|}{0.0} & 0.0* & 0.0 & 0.0 & 0.0 & 1.0 & 31.0* & {\ul 34.0} & 1.0 & \multicolumn{1}{c|}{26.0} & \textbf{38.0} \\
 & ESD & \multicolumn{1}{c|}{0.0} & 0.0 & 0.0* & 0.0 & 0.0 & 0.0 & 3.0 & {\ul 32.0*} & 9.0 & \multicolumn{1}{c|}{3.0} & \textbf{34.0} \\
 & FMN & \multicolumn{1}{c|}{0.0} & 1.0 & 0.0 & 0.0* & 0.0 & 1.0 & 3.0 & 9.0 & {\ul 51.0*} & \multicolumn{1}{c|}{1.0} & \textbf{54.0} \\
 & CA & \multicolumn{1}{c|}{0.0} & 0.0 & 0.0 & 0.0 & 0.0* & 3.0 & 19.0 & \textbf{81.0} & {\ul 60.0} & \multicolumn{1}{c|}{44.0*} & 55.0 \\
\multirow{-5}{*}{\begin{tabular}[c]{@{}c@{}}Van Gogh\\ (Style)\end{tabular}} & \cellcolor[HTML]{FFC3C3}Average & \multicolumn{1}{c|}{\cellcolor[HTML]{FFC3C3}0.0} & \cellcolor[HTML]{FFC3C3}0.3 & \cellcolor[HTML]{FFC3C3}0.0 & \cellcolor[HTML]{FFC3C3}0.0 & \cellcolor[HTML]{FFC3C3}0.0 & \cellcolor[HTML]{FFC3C3}1.3 & \cellcolor[HTML]{FFC3C3}14.0 & \cellcolor[HTML]{FFC3C3}{\ul 39.0} & \cellcolor[HTML]{FFC3C3}30.3 & \multicolumn{1}{c|}{\cellcolor[HTML]{FFC3C3}18.5} & \cellcolor[HTML]{FFC3C3}\textbf{45.3} \\ \hline
 & UCE & \multicolumn{1}{c|}{0.5} & 0.0* & 0.5 & 0.5 & 0.5 & 0.0 & \textbf{70.0*} & {\ul 49.0} & 34.0 & \multicolumn{1}{c|}{33.5} & {\ul 49.0} \\
 & ESD & \multicolumn{1}{c|}{0.0} & 0.0 & 0.5* & 0.0 & 0.0 & 0.0 & 11.5 & {\ul 50.5*} & 37.0 & \multicolumn{1}{c|}{14.5} & \textbf{61.5} \\
 & FMN & \multicolumn{1}{c|}{0.5} & 0.0 & 0.5 & 1.0* & 1.0 & 0.0 & 11.5 & 36.5 & \textbf{92.0*} & \multicolumn{1}{c|}{36.0} & {\ul 76.5} \\
 & CA & \multicolumn{1}{c|}{1.0} & 0.0 & 1.0 & 0.0 & 0.0* & 0.0 & 52.5 & 74.5 & {\ul 79.5} & \multicolumn{1}{c|}{78.0*} & \textbf{86.0} \\
\multirow{-5}{*}{\begin{tabular}[c]{@{}c@{}}Marc\\ Chagall\\ (Style)\end{tabular}} & \cellcolor[HTML]{FFC3C3}Average & \multicolumn{1}{c|}{\cellcolor[HTML]{FFC3C3}0.5} & \cellcolor[HTML]{FFC3C3}0.0 & \cellcolor[HTML]{FFC3C3}0.6 & \cellcolor[HTML]{FFC3C3}0.4 & \cellcolor[HTML]{FFC3C3}0.4 & \cellcolor[HTML]{FFC3C3}0.0 & \cellcolor[HTML]{FFC3C3}36.4 & \cellcolor[HTML]{FFC3C3}52.6 & \cellcolor[HTML]{FFC3C3}{\ul 60.6} & \multicolumn{1}{c|}{\cellcolor[HTML]{FFC3C3}40.5} & \cellcolor[HTML]{FFC3C3}\textbf{68.3} \\ \hline
 & UCE & \multicolumn{1}{c|}{0.0} & 1.4* & 0.7 & 0.0 & 2.1 & 0.0 & {\ul 20.9*} & 3.7 & 1.5 & \multicolumn{1}{c|}{5.2} & \textbf{41.8} \\
 & ESD & \multicolumn{1}{c|}{10.4} & 11.3 & 13.5* & 5.7 & 12.1 & 31.9 & 60.4 & {\ul 70.1*} & 34.3 & \multicolumn{1}{c|}{67.2} & \textbf{72.4} \\
 & FMN & \multicolumn{1}{c|}{56.0} & 79.4 & 75.9 & 68.8* & 68.1 & 61.7 & {\ul 82.8} & 79.1 & 70.9* & \multicolumn{1}{c|}{76.1} & \textbf{87.3} \\
 & CA & \multicolumn{1}{c|}{2.2} & 7.8 & 3.5 & 3.5 & 5.7* & 51.1 & 50.0 & 42.5 & 19.4 & \multicolumn{1}{c|}{\textbf{64.9*}} & {\ul 58.2} \\
\multirow{-5}{*}{\begin{tabular}[c]{@{}c@{}}Nudity\\ (NSFW)\end{tabular}} & \cellcolor[HTML]{FFC3C3}Average & \multicolumn{1}{c|}{\cellcolor[HTML]{FFC3C3}17.2} & \cellcolor[HTML]{FFC3C3}25.0 & \cellcolor[HTML]{FFC3C3}23.4 & \cellcolor[HTML]{FFC3C3}19.5 & \cellcolor[HTML]{FFC3C3}22.0 & \cellcolor[HTML]{FFC3C3}36.2 & \cellcolor[HTML]{FFC3C3}{\ul 53.5} & \cellcolor[HTML]{FFC3C3}48.9 & \cellcolor[HTML]{FFC3C3}31.5 & \multicolumn{1}{c|}{\cellcolor[HTML]{FFC3C3}53.4} & \cellcolor[HTML]{FFC3C3}\textbf{64.9} \\ \hline
 & UCE & \multicolumn{1}{c|}{0.0} & 0.0* & 0.0 & 0.0 & 0.0 & 0.0 & {\ul 56.0*} & \textbf{67.0} & 0.0 & \multicolumn{1}{c|}{10.0} & 39.0 \\
 & ESD & \multicolumn{1}{c|}{0.0} & 0.0 & 0.0* & 0.0 & 0.0 & 0.0 & 0.0 & {\ul 18.0*} & 5.0 & \multicolumn{1}{c|}{0.0} & \textbf{32.0} \\
 & FMN & \multicolumn{1}{c|}{0.0} & 0.0 & 0.0 & 1.0* & 0.0 & 0.0 & 0.0 & 6.0 & {\ul 56.0*} & \multicolumn{1}{c|}{0.0} & \textbf{81.0} \\
 & CA & \multicolumn{1}{c|}{0.0} & 0.0 & 0.0 & 0.0 & 0.0* & 0.0 & 38.0 & {\ul 59.0} & 47.0 & \multicolumn{1}{c|}{41.0*} & \textbf{70.0} \\
\multirow{-5}{*}{\begin{tabular}[c]{@{}c@{}}Barack\\ Obama\\ (ID)\end{tabular}} & \cellcolor[HTML]{FFC3C3}Average & \multicolumn{1}{c|}{\cellcolor[HTML]{FFC3C3}0.0} & \cellcolor[HTML]{FFC3C3}0.0 & \cellcolor[HTML]{FFC3C3}0.0 & \cellcolor[HTML]{FFC3C3}0.3 & \cellcolor[HTML]{FFC3C3}0.0 & \cellcolor[HTML]{FFC3C3}0.0 & \cellcolor[HTML]{FFC3C3}23.5 & \cellcolor[HTML]{FFC3C3}{\ul 37.5} & \cellcolor[HTML]{FFC3C3}27.0 & \multicolumn{1}{c|}{\cellcolor[HTML]{FFC3C3}12.8} & \cellcolor[HTML]{FFC3C3}\textbf{55.5} \\ \hline
 & UCE & \multicolumn{1}{c|}{0.0} & 0.0* & 0.0 & 0.0 & 0.0 & 0.0 & {\ul 57.5*} & 49.0 & 5.0 & \multicolumn{1}{c|}{\textbf{86.5}} & 56.5 \\
 & ESD & \multicolumn{1}{c|}{0.0} & 0.0 & 0.0* & 0.0 & 0.0 & 0.0 & 3.5 & {\ul 61.0*} & 15.5 & \multicolumn{1}{c|}{25.5} & \textbf{69.0} \\
 & FMN & \multicolumn{1}{c|}{3.0} & 3.0 & 1.0 & 2.0* & 0.0 & 1.0 & 0.5 & 55.5 & \textbf{85.0*} & \multicolumn{1}{c|}{45.5} & {\ul 70.5} \\
 & CA & \multicolumn{1}{c|}{0.0} & 1.0 & 0.0 & 0.0 & 0.0* & 0.0 & 36.5 & 50.5 & 54.5 & \multicolumn{1}{c|}{\textbf{93.0*}} & {\ul 69.5} \\
\multirow{-5}{*}{\begin{tabular}[c]{@{}c@{}}Emma\\ Watson\\ (ID)\end{tabular}} & \cellcolor[HTML]{FFC3C3}Average & \multicolumn{1}{c|}{\cellcolor[HTML]{FFC3C3}0.8} & \cellcolor[HTML]{FFC3C3}1.0 & \cellcolor[HTML]{FFC3C3}0.3 & \cellcolor[HTML]{FFC3C3}0.5 & \cellcolor[HTML]{FFC3C3}0.0 & \cellcolor[HTML]{FFC3C3}0.3 & \cellcolor[HTML]{FFC3C3}24.5 & \cellcolor[HTML]{FFC3C3}54.0 & \cellcolor[HTML]{FFC3C3}40.0 & \multicolumn{1}{c|}{\cellcolor[HTML]{FFC3C3}{\ul 62.6}} & \cellcolor[HTML]{FFC3C3}\textbf{66.4} \\ \Xhline{1.0pt}
\end{tabular}}
\end{table*}

\end{document}